\documentclass[10pt,twocolumn,letterpaper]{article}

\usepackage[pagenumbers]{cvpr} %

\usepackage{graphicx}
\usepackage{amsmath}
\usepackage{amssymb}
\usepackage{booktabs}

\usepackage{amsmath,amsfonts,bm}

\def\eqref#1{equation~\ref{#1}}

\def\1{\bm{1}}

\DeclareMathAlphabet{\mathsfit}{\encodingdefault}{\sfdefault}{m}{sl}
\SetMathAlphabet{\mathsfit}{bold}{\encodingdefault}{\sfdefault}{bx}{n}

\newcommand{\E}{\mathbb{E}}

\newcommand{\Var}{\mathrm{Var}}

\newcommand{\w}{{\bf W}}
\newcommand{\wa}{{{\bf W}^{1}}}
\newcommand{\wb}{{{\bf W}^{2}}}
\newcommand{\y}{{{\bf y}}}
\newcommand{\x}{{{\bf x}}}

\newcommand{\sio}{{\bf \Sigma}^{yx}}

\newcommand{\Ub}{{{\bf U}}}
\newcommand{\A}{{{\bf A}}}

\newcommand{\Vt}{{{\bf V}^{T}}}

\newcommand{\mk}[1]{}
\newcommand{\alessandro}[1]{}
\newcommand{\note}[1]{}

\usepackage[pagebackref,breaklinks,colorlinks]{hyperref}

\usepackage[capitalize]{cleveref}
\crefname{section}{Sec.}{Secs.}
\Crefname{section}{Section}{Sections}
\Crefname{table}{Table}{Tables}
\crefname{table}{Tab.}{Tabs.}

\def\cvprPaperID{12009} %
\def\confName{CVPR}
\def\confYear{2023}

\begin{document}

\title{
Critical Learning Periods for Multisensory Integration in Deep Networks
}

\author{Michael Kleinman$^1$\thanks{Work conducted during an internship at AWS AI Labs.} \quad Alessandro Achille$^2$ \quad Stefano Soatto$^{2}$\\
$^1$University of California, Los Angeles \quad $^2$AWS AI Labs\\
\texttt{michael.kleinman@ucla.edu}\ \  \texttt{\{aachille,soattos\}@amazon.com} 
}

\maketitle

\begin{abstract}
We show that the ability of a neural network to integrate information from diverse sources hinges critically on being exposed to properly correlated signals during the early phases of training. Interfering with the learning process during this initial stage can permanently impair the development of a skill, both in artificial and biological systems where the phenomenon is known as a \emph{critical learning period}. We show that critical periods arise from the complex and unstable early transient dynamics, which are decisive of final performance of the trained system and their learned representations. This evidence challenges the view, engendered by analysis of wide and shallow networks, that early learning dynamics of neural networks are simple, akin to those of a linear model. Indeed, we show that even deep linear networks exhibit critical learning periods for multi-source integration, while shallow networks do not.  To better understand how the internal representations change according to disturbances or sensory deficits, we introduce a new measure of source sensitivity, which allows us to track the inhibition and integration of sources during training.
Our analysis of inhibition suggests cross-source reconstruction as a natural auxiliary training objective, and indeed we show that architectures trained with cross-sensor reconstruction objectives are remarkably more resilient to critical periods.
Our findings suggest that the recent success in self-supervised multi-modal training compared to previous supervised efforts may be in part due to more robust learning dynamics and not solely due to better architectures and/or more data. 
\end{abstract}

\section{Introduction}

Learning generally benefits from exposure to diverse sources of information, including different sensory modalities, views, or features. Multiple sources can be more informative than the sum of their parts. For instance, both views of a random-dot stereogram  are needed to extract the \textit{synergistic information}, which is absent in each individual view  \cite{julesz1960binocular}. More generally, multiple sources can help identify latent common factors of variation relevant to the task, and separate them from source-specific nuisance variability, as done in contrastive learning.

Much information fusion work in Deep Learning focuses on the design of the architecture, as different sources may require different architectural biases to be efficiently encoded. We instead focus on the \textit{learning dynamics}, since effective fusion of different sources relies on complex phenomena beginning during the early epochs of training. In fact, even slight interference with the learning process during this \textit{critical period} can permanently damage a network's ability to harvest synergistic information. Even in animals, which excel at multi-sensor fusion, a temporary deficit in one source during early development can permanently impair the learning process: congenital strabismus in humans can cause permanent loss of stereopsis if not corrected sufficiently early; similarly, visual/auditory misalignment can impair the ability of barn owls to localize prey \cite{kandel2013principles}. 
In artificial networks, the challenge of integrating different sources has been noted in visual question answering (VQA), where the model often resorts to encoding less rich but more readily accessible textual information \cite{agrawal2016analyzing, cadene2019rubi}, ignoring the visual modality, or in audio-visual processing, where acoustic information is often washed out by visual information \cite{wang2020makes}.

Such failures are commonly attributed to  the mismatch in learning speed between sources, or their ``information asymmetry'' for the task. It has also been suggested, based on limiting analysis for wide networks, that the initial dynamics of DNNs are very simple \cite{hu2020surprising}, seemingly in contrast with evidence from biology.  In this paper, we instead argue that \textit{the early learning dynamics of information fusion in deep networks are both highly complex and brittle, to the point of exhibiting critical learning periods similar to biological systems.}

\begin{figure*}
    \centering
    \includegraphics[width=\textwidth]{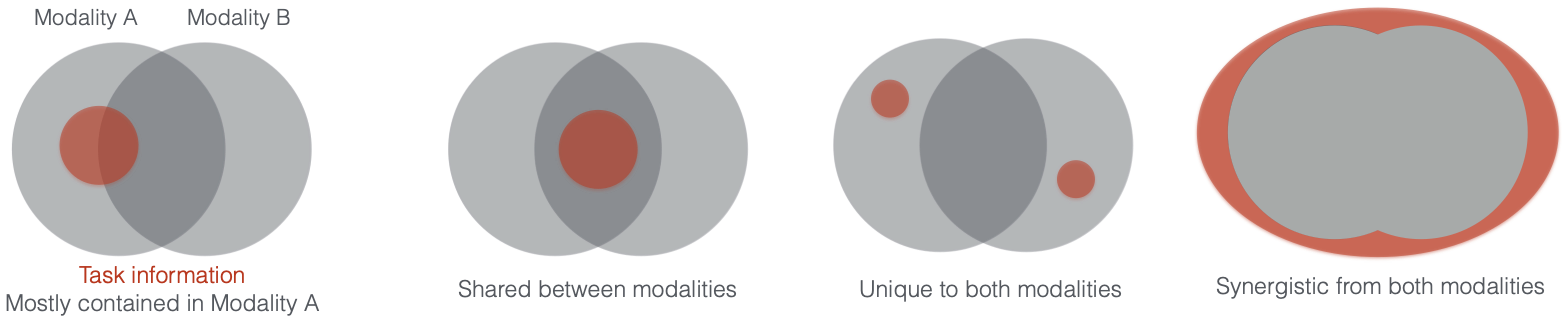}
    \vspace{-1em}
    \caption{\textbf{Decomposition of information between different modalities.} Two modalities can have unique information, common information (denoted by the overlap in the venn-diagram), or synergistic information (denoted by the additional ellipse in the right panel). Task-relevant information (shown in red) can be distributed in a variety of ways across the different modalities. Task-relevant information can be mostly present in Modality A (left), shared between modalities (center-left), or could require unique (center-right) or synergistic information from both modalities (right).
    \vspace{-1em}
    }
    \label{fig:overview_schem}
\end{figure*}

In Sect.~\ref{sec:theory}, we show that shallow networks do not exhibit critical periods when learning to fuse diverse sources of information, but \textit{deep} networks do. Even though, unlike animals,  artificial networks do not age, their learning success is still decided during the early phases of training. 
The existence of critical learning periods for information fusion is not an artifact of annealing the learning rate or other details of the optimizer and the architecture. In fact, we show that critical periods for fusing information are present even in a simple deep linear network. This contradicts the idea that deep networks exhibit trivial early dynamics \cite{hu2020surprising,lee2019wide}. We provide an interpretation for critical periods in linear networks in terms of mutual inhibition/reinforcement between sources, manifest through sharp transitions in the learning dynamics, which in turn are related to the intrinsic structure of the underlying data distribution.

In Sect.~\ref{sec:sensitivity}, we introduce a metric called ``Relative Source Variance'' to quantify the dependence of units in a representation to individual sources, allowing us to better understand inhibition and fusion between sources. Through it, in Sect.~\ref{sec:empirical},
we show that temporarily reducing the information in one source, or breaking the correlation between sources, can permanently change the overall amount of  information in the learned representation. Moreover, even when downstream performance is not significantly affected, such temporary changes result in units that are highly polarized and process only information from one source or the other. 
Surprisingly, we found that the final representations in our artificial networks that were exposed to a temporary deficit mirrored single-unit animal representations exposed to analogous deficits (Fig.~\ref{fig:blur_deficit}, Fig.~\ref{fig:strabismus}).

We hypothesize that features inhibit each other because they are competing to solve the task. But if the competitive effect is reduced, such as through an auxiliary cross-source reconstruction task, the different sources can interact synergistically. This supports cross-modal reconstruction as a practical self-supervision criterion.
In Sect.~\ref{sec:video-unsupervised}, we show that indeed auxiliary cross-source reconstruction can stabilize the learning dynamics and prevent critical periods.  This lends an alternate interpretation for the recent achievements in multi-modal learning as due to the improved stability of the early learning dynamics due to auxiliary cross-modal reconstruction tasks, rather than to the design of the architecture.

\begin{figure*}[t!]
    \centering
    \raisebox{0.5cm}{\includegraphics[width=0.3\linewidth]{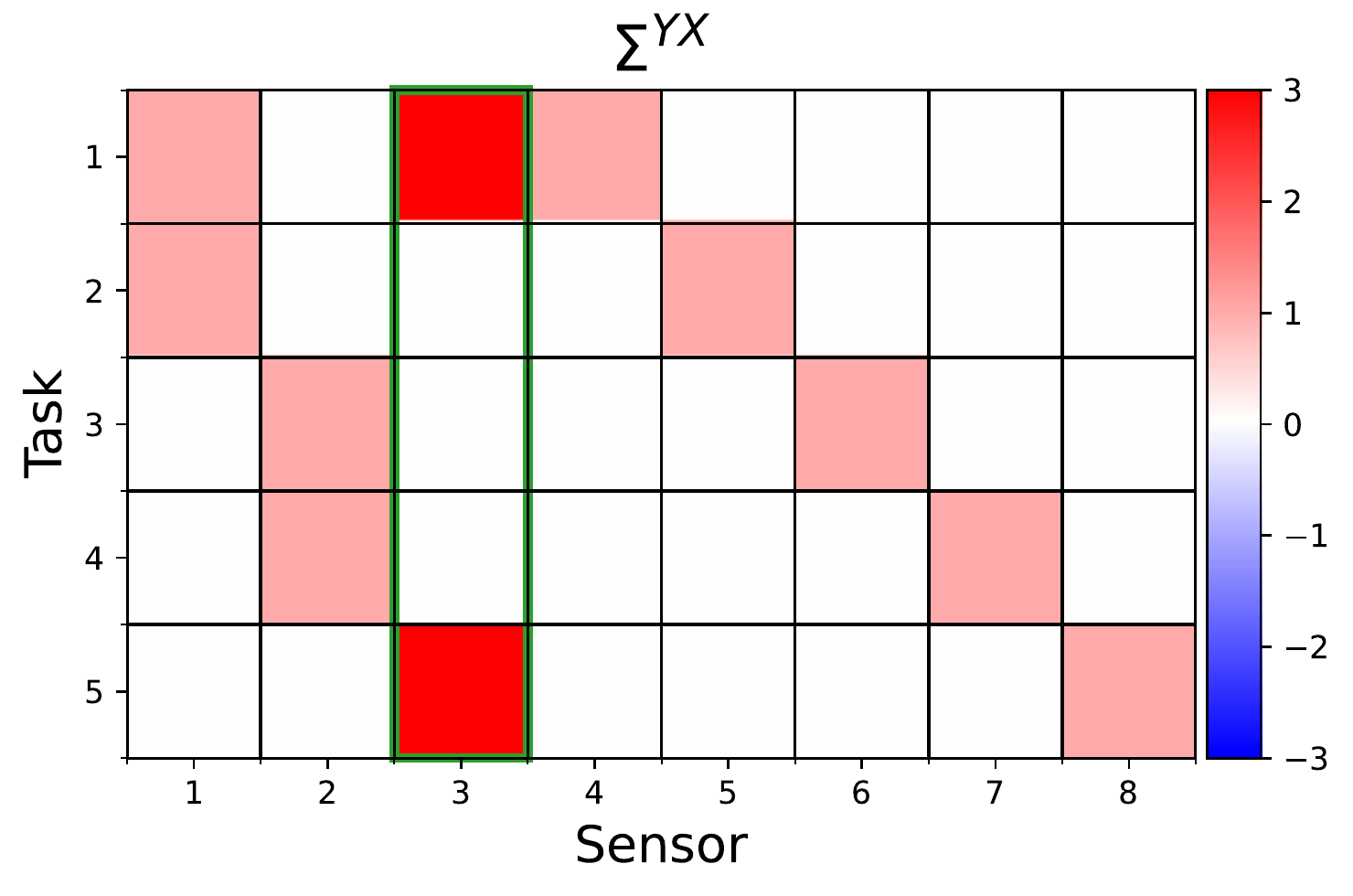}}
    \includegraphics[width=0.25\linewidth]{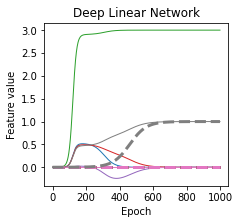}
    \includegraphics[width=0.25\linewidth]{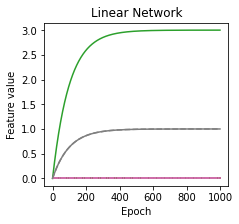}
    \vspace{-1.2em}
    \caption{
    \textbf{(Left)} $\Sigma^{yx}$, with the highlighted green column representing the sensor that was dropped.
    \textbf{(Center)} 
    We show total weights attributed to each feature (shown in different colors) during training in a deep linear network.
    The solid lines represent the dynamics when training with all features. The dashed lines represent the behavior when training with the green feature disabled. Note that disabling the green feature prevents the gray feature from being learned during the initial transient
    \textbf{(Right)} Same experiment with a shallow linear network. In this case the learning dynamics of the gray feature perfectly overlap in both cases. 
    \vspace{-1em}
    }
    \label{fig:deep_linear}
\end{figure*}

Empirically, we show the existence of critical learning periods for multi-source integration using state-of-the-art architectures (Sect.~\ref{sec:video-supervised}-\ref{sec:video-unsupervised}). To isolate different factors that may contribute to low-performance on multi-modal tasks (mismatched training dynamics, different informativeness), we focus on tasks where the sources of information are symmetric and homogeneous, in particular stereo and multi-view imagery. Even in this highly controlled setting, we observe the effect of critical periods both in downstream performance and/or in unit polarization. 
Our analysis suggests that pre-training on one modality, for instance text, and then adding additional pre-trained backbones, for instance visual and acoustic, as advocated in recent trends with Foundation Models, yields representations that fail to encode synergistic information. Instead, training should be performed across modalities at the outset.  
Our work also suggests that asymptotic analysis is irrelevant for deep network fusion, as their fate is sealed during the initial transient learning. Also, conclusions drawn from wide and shallow networks do not transfer to deep networks in use in practice.

\subsection{Related Work}
\label{sec:related_work}

\textbf{Multi-sensor learning.}
There is a large literature on sensor fusion in early development \cite{smith2005development}, including homogeneous sensors that are spatially dislocated (e.g., two eyes), or time-separated  (e.g., motion), and heterogeneous sources (e.g., optical and acoustic, or visual and tactile). Indeed, given \emph{normal learning}, humans and other animals have the remarkable ability to integrate multi-sensory data, such as incoming visual stimuli coming into two eyes, as well as corresponding haptic and audio stimuli. Monkeys have been shown to be adept at combining and leveraging arbitrary sensory feedback information \cite{dadarlat2015learning}.

In deep learning,
multi-modal (or \emph{multi-view} learning) learning typically falls into two broad categories: learning a joint representation (fusion of information) and learning an aligned representation (leveraging coordinated information in the multiple views) \cite{baltruvsaitis2018multimodal}. A fusion-based approach is beneficial if there is synergistic information available in the different views, while an alignment-based approach is helpful is there is shared information common to the different views (Fig.~\ref{fig:overview_schem}). Such a division of information typically affects architectural and model choices: synergistic information requires the information from the different modalities to be fused or combined, whereas shared information often serves as a self-supervised signal that can align information from the different modalities, as in contrastive learning \cite{tian2020contrastive, tian2020makes, chen2020simple}, correlation based \cite{andrew2013deep}, and information-theoretic approaches \cite{kleinman2022gacs, kleinman2021redundant}.

\textbf{Critical periods in animals and deep networks:} Such architectural considerations often neglect the impact coming from multisensory learning dynamics, where information can be learned at different speeds from each sensor \cite{wu2022characterizing}.
Indeed, \cite{wiesel1982postnatal} showed that humans and animals are peculiarly sensitive to changes in the distribution of sensory information early in training, in a phenomenon known as \textit{critical periods}. Critical periods have since been described in many different species and sensory organs. For example, barn owls originally exposed to misaligned auditory and visual information cannot properly localize prey \cite{knudsen1990sensitive}.
Somewhat surprisingly, similar critical periods for learning have also been observed in deep networks.
\cite{achille2018critical} found that early periods of training were critical for determining the asymptotic network behavior. Additionally, it was found that the timing of regularization was important for determining asymptotic performance \cite{golatkar2019regularization}, with regularization during the initial stages of training having the most influential effect. 

\textbf{Masked/de-noising Autoencoders:}
Reconstructing an input from a noisy or partial observation has been long used as a form of supervision. Recently, an in part due the successful usage of transformers in language  \cite{vaswani2017attention} and vision tasks \cite{dosovitskiy2020image}, such a pre-training strategy has been successfully applied to text \cite{devlin2018bert} and vision tasks \cite{he2021masked}.  An extension of this has been recently applied to multi-modal data \cite{bachmann2022multimae}.

\textbf{Models of learning dynamics} 
We consider two approaches to gain analytic insight into the learning dynamics of deep networks. \cite{saxe2013exact, saxe2019mathematical} assume that the input-output mapping is done by a deep linear network. We show that under this model critical periods may exist. \cite{hu2020surprising, lee2019wide} assume instead  infinitely wide networks, resulting in a model linear with respect to the parameters. In this latter case, no critical period is predicted contradicting our empirical observations on finite networks.

\begin{figure*}[t!]
    \centering
    \includegraphics[width=\textwidth]{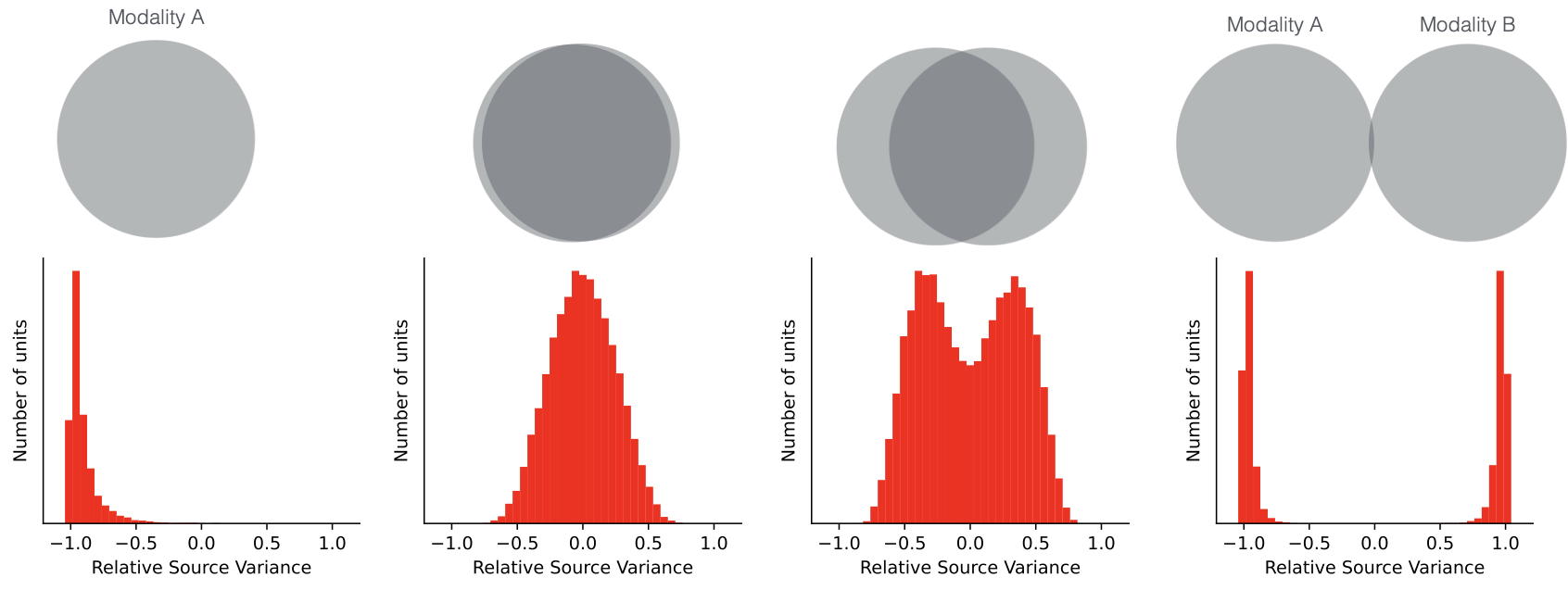}
    \caption{\textbf{Example RSV distributions and relation to information diagrams.} \textbf{(Left)} Representations that vary predominantly due to one modality. \textbf{(Center-Left, Center-right)} All units in the representation vary nearly equally with both modalities. \textbf{(Right)} Units in the representation that vary uniquely with each sensor, which is reflected by a polarized RSV distribution. 
    \vspace{-1em}
    }
    \label{fig:fsv_schem}
\end{figure*}

\section{A model for critical periods in sensor-fusion}
\label{sec:theory}

We want to establish what is the difference, in terms of learning dynamics, between learning how to use two sources of information at the same time, or learning how to solve a task using each modality separately and then merging the results.
In particular we consider the counterfactual question: if we disable sensor A during training, would this change how we learn to use sensor B?
To start, let's consider the simple case of a linear regression model $\y = \w \x$ trained with a mean square error loss
    \[
    L = \frac{1}{N} \sum_{i=1}^N \frac{1}{2}||\y^{(i)} - \w \x^{(i)}||^2
    \]
    where $D = \{(\x^{(i)}, \y^{(i)})\}_{i=1}^N$ is a training set of i.i.d. samples. In this simplified setting, we consider each component $x_k$ of $\x$ as coming from a different sensor or source.
To simplify even further, we assume that the inputs have been whitened, so that the input correlation matrix ${\bf{\Sigma}}^x=\frac{1}{N} \sum_i \x^{(i)} {\x^{{(i)}T}} = \mathbf{I}$.

In this case, the learning dynamics of any source is independent from the others. In fact, the gradient of the weight $w_{jk}$ associated to $x_k$ and $y_j$ is given by
\[- \nabla_{w_{jk}} L(\w) = - \nabla_{w_{jk}} \frac{1}{N} \sum_{i=1}^N \frac{1}{2}||\y^{(i)} - \w \x^{(i)}||^2 = {{\Sigma}}_{jk}^{yx} - w_{jk} \]
and does not depend on any $w_{hl}$ with $w_{hl} \neq w_{jk}$. The answer to the counterfactual question is thus negative in this setting: adding or removing one source of information (or output) will not change how the model learns to extract information from the other sources.
However, we now show that the addition of depth, even \textit{without} taking introducing non-linearities, makes the situation radically different.

To this effect, consider a deep linear network with one hidden layer $\y= \wb \wa \x$. This network has the same expressive power (and the same global optimum) as the previous model. However, this introduces a mutual dependency between sensors (due to the shared layer) that can ultimately lead to critical periods in cross-sensor learning.
To see this, we use an analytical expression of the learning dynamics for two-layer deep networks \cite{saxe2013exact, saxe2019mathematical}.
Let $\sio = \frac{1}{N}\sum_{i=1}^N \y^{(i)} \x^{{(i)}T}$ be the cross-correlation matrix between the inputs $\x$ and the target vector $\y$\footnote{Note that $\w=\sio$ is also the global minimum of the MSE loss $L = \frac{1}{N} \sum_i \frac{1}{2} ||\y^{(i)} - \w \x^{(i)}||^2$.} and let $\sio = U S V^T$ be its singular-value decomposition (SVD). \cite{saxe2019mathematical} shows that the total weight $\w(t) =\wb(t) \wa(t)$ assigned to each source at time $t$ during the training can be written as
\begin{align}
 \w(t) &= \wb(t)\wa(t) = \Ub \A(t) \Vt \\ 
 &= \sum_{\alpha} a_\alpha(t) \mathbf{u}^\alpha \mathbf{v}^{\alpha T}   
\end{align}
where
\begin{equation}
 a_\alpha(t) = \frac{s_\alpha e^{2 s_\alpha t/ \tau}}{e^{2 s_\alpha t/\tau} - 1 + s_\alpha/ a_\alpha^0}.     
\end{equation}

This leads to non-linear learning dynamics where different features are learned at sharply distinct points in time \cite{saxe2019mathematical}. Moreover, it leads to entanglement between the learning dynamics of different sources due to the eigenvectors $\mathbf{v}^{\alpha}$ mixing multiple sources. 

Disabling (or adding) a source of information corresponds to removing (or adding) a column to the matrix $\sio$, which in turns affects its singular-value decomposition and the corresponding learning dynamics.
To see how this change may affect the learning dynamics, in Fig.~\ref{fig:deep_linear} we compare the weights associated to each sensor during training for one particular task. In solid we show the dynamics with all sensors active at the same time. In dashed line we show the dynamics when one of the sensor is disabled.
We see that disabling a sensor (green in the figure) can completely inhibit learning of other task-relevant features (e.g., the gray feature) during the initial transient.
This should be compared with the learning dynamics of a shallow one-layer network (Fig.~\ref{fig:deep_linear}, right) where all task-relevant features are learned at the same time, and where removal of a source does not affect the others.

In deep linear networks, the suboptimal configuration learned during the initial transient is eventually discarded, and the network reverts to the globally optimal solution. In the following we show this is not the case for standard non-linear deep networks. While the initial non-trivial interaction between sources of information remain, the non-linear networks are unable to unlearn the suboptimal configurations learned at the beginning (owing to the highly non-convex landscape). This can result in permanent impairments if a source of information is removed during the initial transient of learning, which reflects the trends observed in critical periods in animals.

\begin{figure*}
    \centering
    \includegraphics[width=0.9\textwidth]{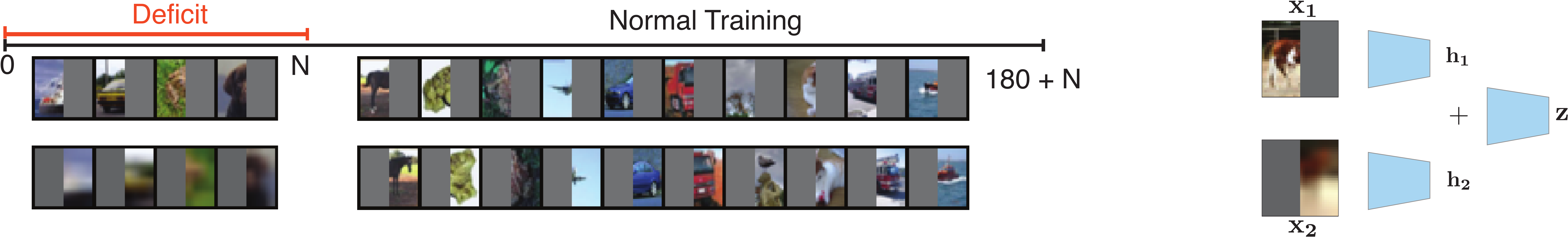} 
    \\
    \includegraphics[width=0.23\textwidth]{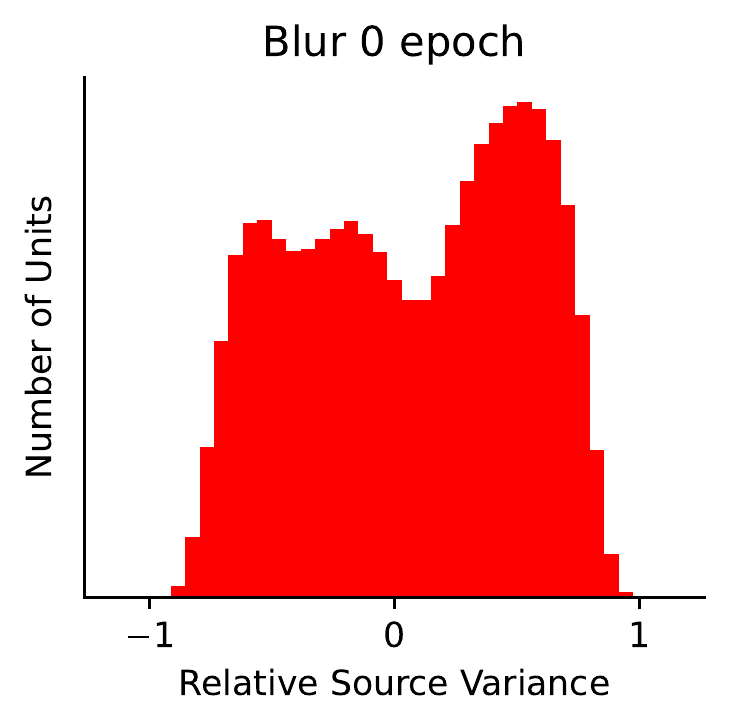}
    \includegraphics[width=0.23\textwidth]{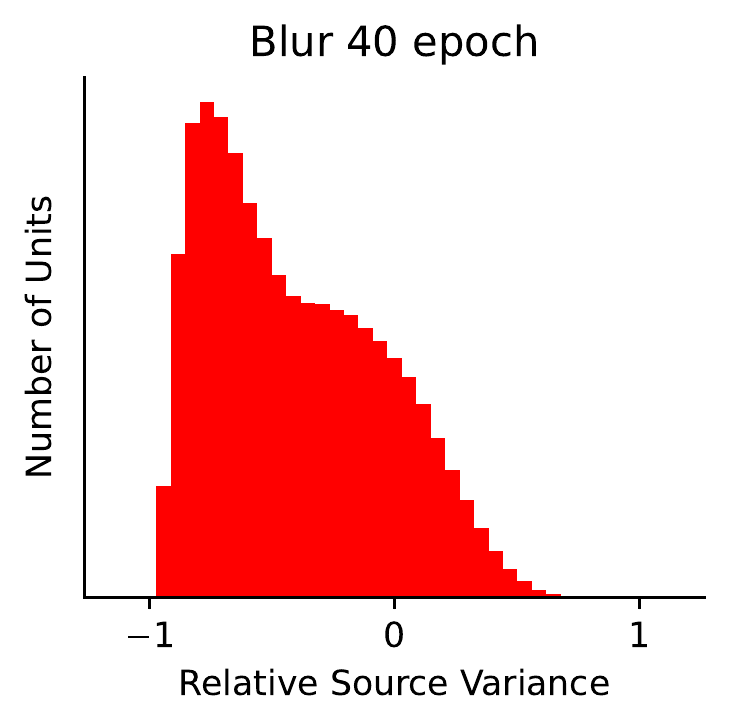}
    \includegraphics[width=0.23\textwidth]{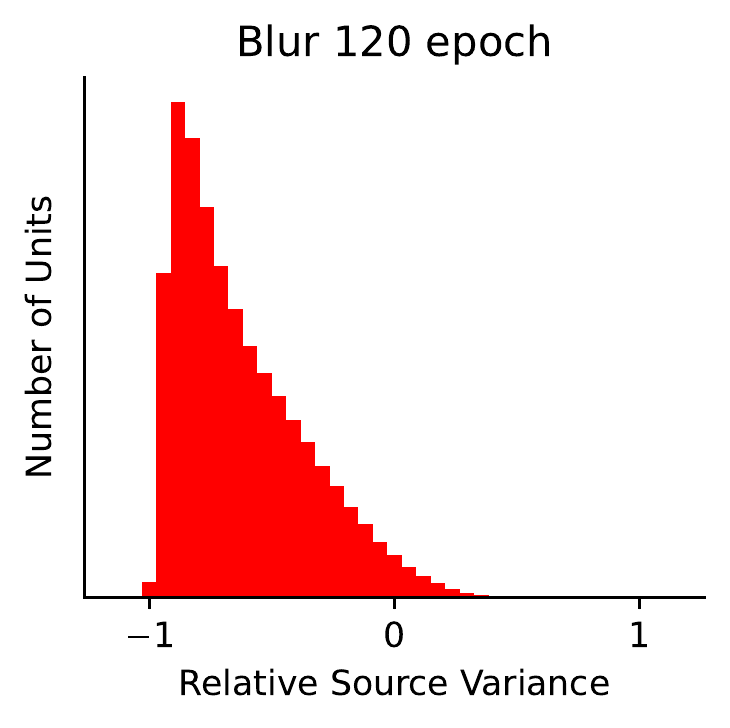}
    \includegraphics[width=0.23\textwidth]{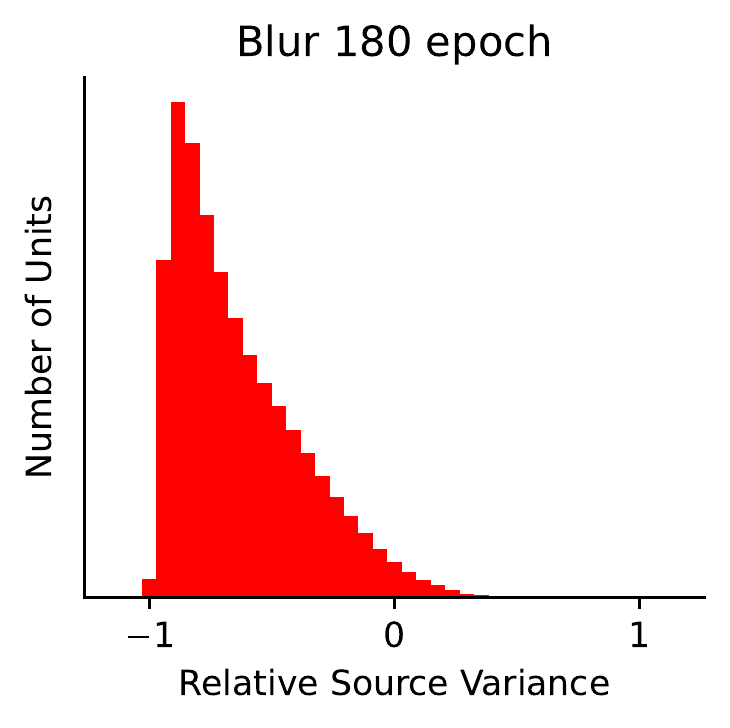}
    \vspace{-1em}
    \caption{ \textbf{Experimental setup and sensor selectivity as a function of a blurring deficit length.}
     \textbf{(Top)} In our experiments, we train the network with a deficit (blurred images to one pathway shown here) for the first $N$ epochs, and then continue training with normal images for 180 more epochs. We feed each half of an image to the early stages of a ResNet-18, and then additively combine the representations from both pathways (followed by stages of common processing). We refer to this architecture as \emph{Split-ResNet}.
     \textbf{(Bottom)} RSV distribution of units in last layer representation $z$ for increasing duration of deficit (blur to one pathway) after resumption of normal training. With a sufficiently long deficit, the units in the representation remain only sensitive to the initially uncorrupted pathway, and do not vary with the initially corrupted pathway.
     \vspace{-1em}
     }  %
    \label{fig:blur_deficit}
\end{figure*}

\section{Single Neuron Sensitivity Analysis}
\label{sec:sensitivity}

Before studying the empirical behavior of real networks on multi-sensor tasks, we should consider how to quantify the effect of a deficit on a down-stream task. One way is to look at the final performance of the model on the task. For example, animals reared with a monocular deprivation deficit have reduced accuracy on a visual acuity test and, similarly, deep networks may show reduced classification accuracy \cite{achille2018critical}. However, in some cases deficits may not drastically impair the accuracy but may still affect how the model is organized internally. Individuals with strabismus or ambliopia can perform just as well on most tasks, since the individual information coming from each sensor separately is enough to compensate. But the connectivity scheme of the synapses may change so that neurons eventually process only information from one sensor or the other, and not from both together, as observed in individuals without deficits \cite{wiesel1982postnatal}.

To understand whether units in a representation of multisensory inputs depend on both sensors or only a particular sensor, we introduce a measure of \emph{Relative Source Variance}. We first define the \emph{Source Variance} (SV) for unit $i$ of a representation due to sensor A, conditioned on an example $b$ as 
\begin{equation}
SV_i(A, b) = \Var(f(A,B)_i|B = b),     
\end{equation}
where $f$ denotes the mapping from multisensory inputs to the representation and $i$ indexes the unit of the representation. We note that the value of $SV_i(A, b)$ depends on the example $b$. We use an analogous formula for $SV_i(B, a)$. 

\begin{figure*}
    \centering
    \includegraphics[width=0.28\textwidth]{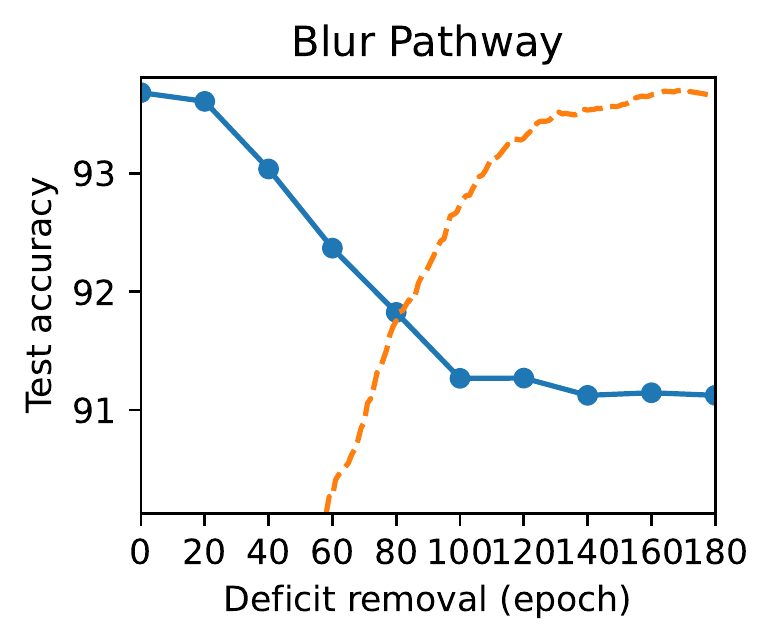}
    \includegraphics[width=0.34\linewidth]{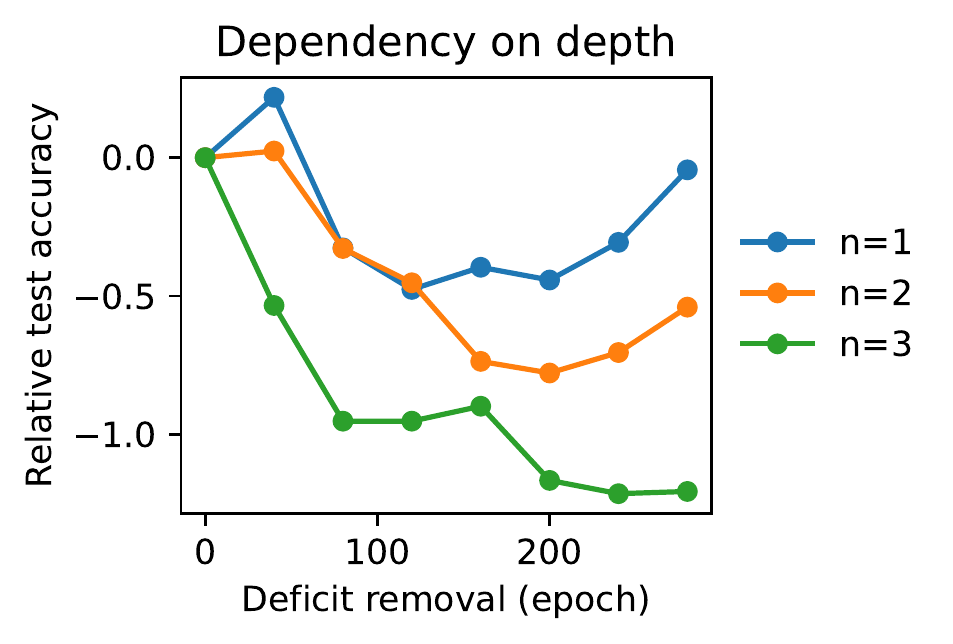}
    \hspace{1em}
    \includegraphics[width=0.265\textwidth]{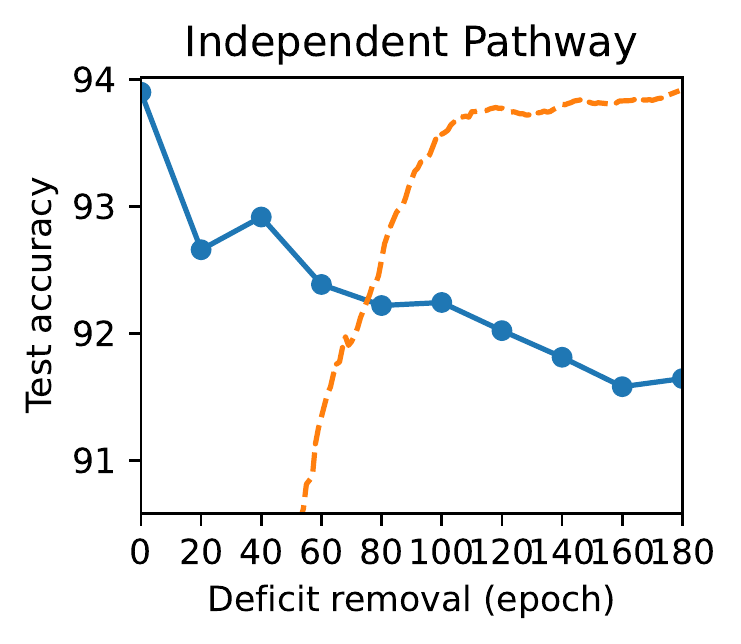}
    \vspace{-1em}
    \caption{\textbf{Decrease in downstream performance as a function of the deficit length.} \textbf{(Left)} Final test accuracy (blue) when applying a blurring deficit to one pathway of Split-ResNet. Even though the network is exposed to a subsequent number of uncorrupted paired observations, the network cannot later learn to optimally fuse the information. The orange dashed line represents accuracy of a normal network during training.
    \textbf{(Center)} The effect of a deficit is most pronounced when increasing the depth of the network (see Appendix for architecture detail).
    \textbf{(Right)} We also observe a degradation of test performance using a dissociation deficit (feeding uncorrelated views). We note that the effect is less marked than the blurring, due to better ability to compensate. Additional runs are shown in Fig.~\ref{fig:robustness}.
    \vspace{-1em}
    }
    \label{fig:depth}
\end{figure*}

\begin{figure*}
    \centering
    \includegraphics[width=0.23\textwidth]{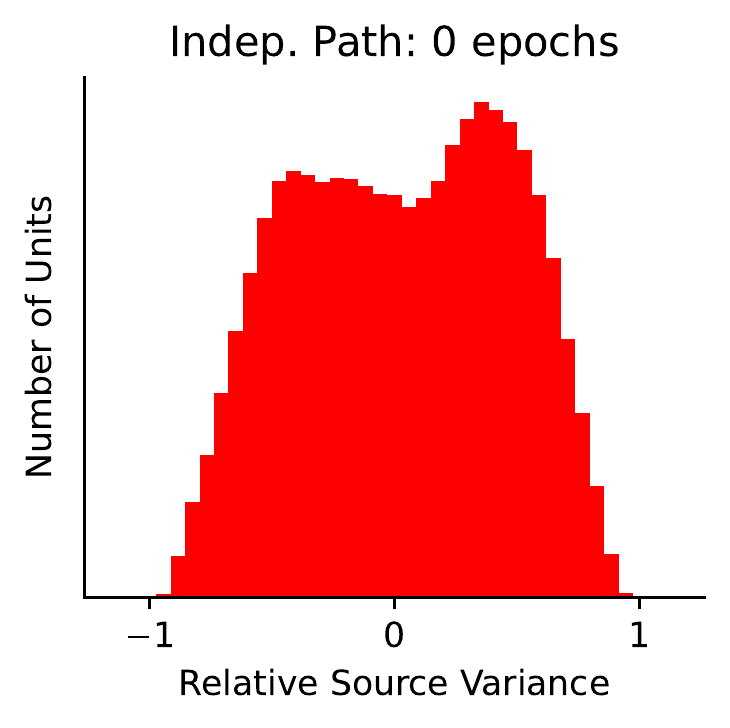}
    \includegraphics[width=0.23\textwidth]{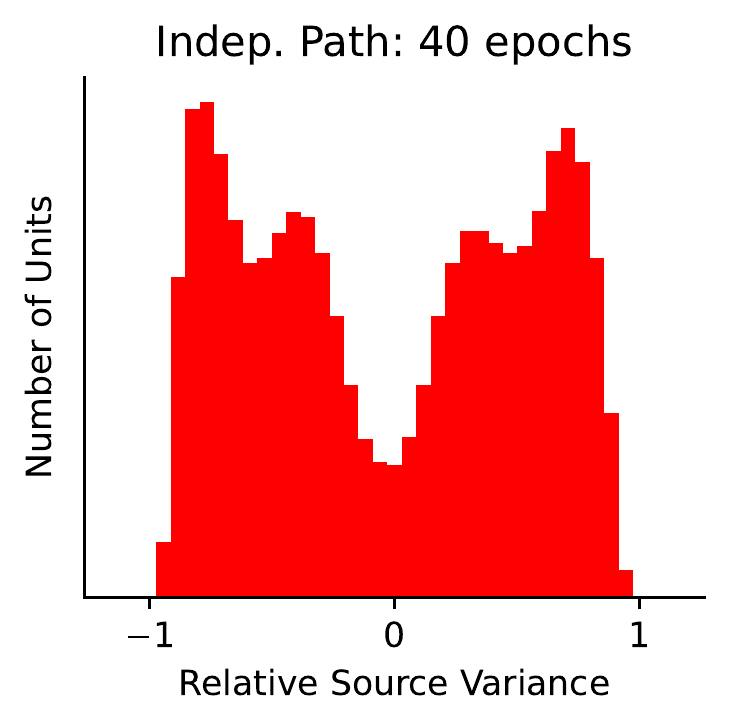}
    \includegraphics[width=0.23\textwidth]{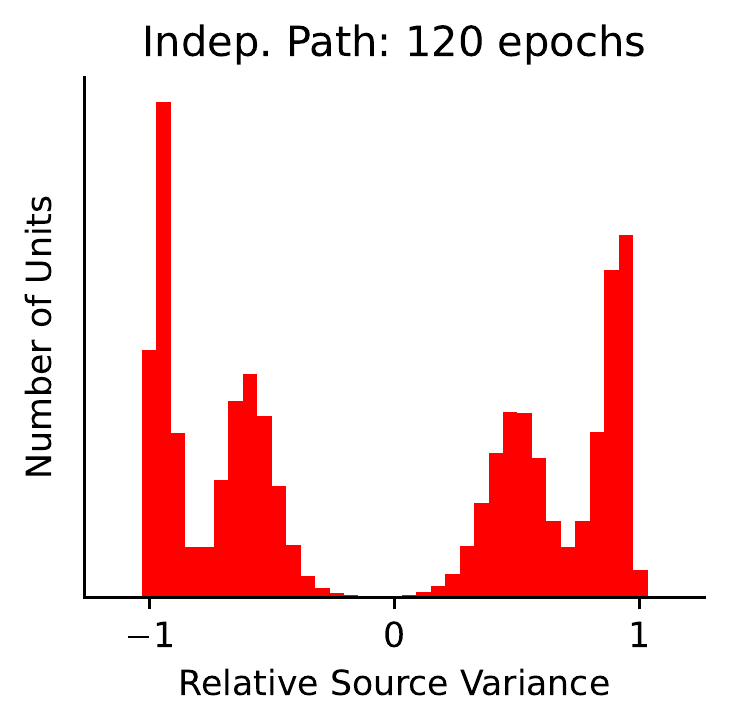}
    \includegraphics[width=0.23\textwidth]{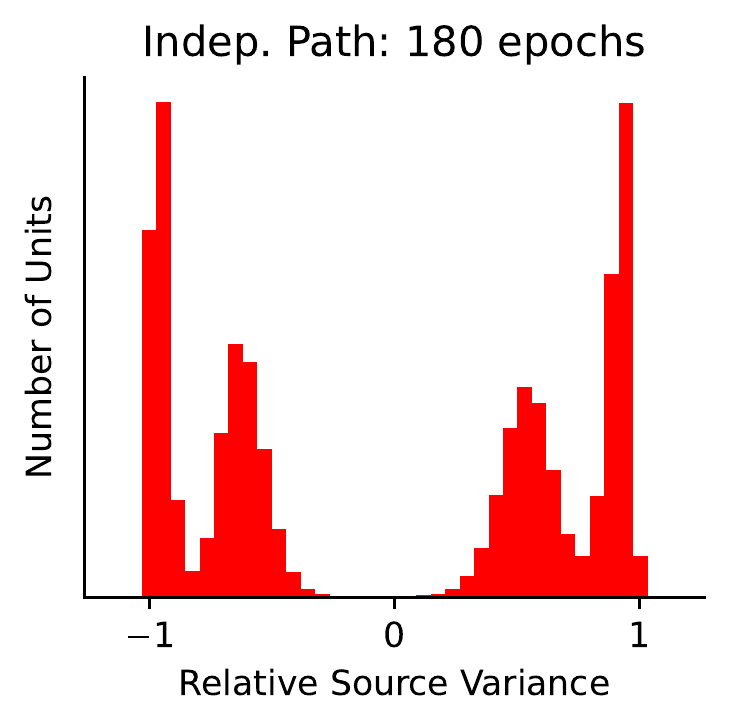}
    \vspace{-1em}
    \caption{ \textbf{Sensor selectivity as a function of a dissociation deficit length.} We examined the asymptotic representations and found that, when exposed to a sufficiently long deficit of broken correlations between the views, the network could no longer learn a bimodal distribution that learned common features, but instead resulted in a polarized representation in which units are sensitive to either view (but none to both). %
    \vspace{-1em}
    }
    \label{fig:strabismus}
\end{figure*}

Typically, we are interested in the distribution of the Source Variance of the units $i$ in a representation, as a function of many examples $a$ and $b$. To capture this, we define a notion of Relative Source Variance (RSV) for unit $i$ as:
\begin{equation}
\label{eq:fsv}
    RSV_i(a,b) = \frac{SV_i(A,b) - SV_i(B,a)}{SV_i(A,b) + SV_i(B,a)}
\end{equation}
If the RSV is $1$, this means that the unit is only sensitive to sensor A, and if the RSV is $-1$, the unit is sensitive to sensor $B$. 
To compute $SV(A,b)$ (and analogously for $SV(B,a)$) from samples, we fix a sample $b$, and vary the inputs $a$, sampling from $a \sim p(a)$. We run this for multiple fixed samples from $b$, performing the computation over a batch. We perform analogous computations for $SV(B,a)$
We compute the $RSV_i(a,b)$ for all units $i$ from a representation, and for many examples $a$ and $b$. We then plot the distribution of RSVs, aggregating across all units (see, e.g., Fig.~\ref{fig:blur_deficit}-\ref{fig:strabismus}). In particular, we track how the distribution changes as a result of sensory deficits and perturbations, as well as how the distribution changes during normal training. Note that $-1 \leq RSV_i(a,b) \leq 1$. If $RSV_i(a,b)=1$ (or -1) is $1$, this means that the unit is only sensitive to sensor $A$ (or $B$).  If $RSV_i(a,b) = 0$ the unit is equally sensitive to both sensors. For controlled simulations (See Appendix \ref{sec:rsv_simulation}), we show the variety of distributions of units in a representation that the RSV can measure in Fig.~\ref{fig:fsv_schem}.
In our experiments, described next, we computed the RSV on units from the final layer before linear classification.

\section{Critical learning periods in deep multi-sensor networks}
\label{sec:empirical}

In this section, we investigate the learning dynamics of deep networks during the initial learning transient when multiple source of information are present. We evaluate how temporary perturbations of the relation between the two sensors during the training can change the final outcome. To exclude possible confounding factors, in all our experiments, the two input sources are perfectly symmetrical (same data distribution and same informativeness for the task) which ensures that any asymmetry observed in the final model is due to the perturbation.

\subsection{Inhibition of a weak source}

Uncorrected vision problems in an eye during early childhood can cause permanent visual impairment in humans, whereas even after correction the patient only sees through the unaffected eye and does not recover vision in the affected eye (ambliopia, or lazy-eye). We explore whether such inhibition of a sensor can happen in DNNs following a similar experimental setup to \cite{achille2018critical}.
To simulate binocular data from single images, we partition each image in a left and right crop and feed each to two separate pathways of the network, which are then fused in an additive manner at a later stage. For each initial pathway, we
used the early stages of a ResNet-18 backbone.
We then simulate the blurry vision of a weak eye by downsampling the input of the right pathway by $4 \times$, and then resized the image to the original size.
After training for $t_0$ initial epochs with the blur deficit, we remove it and train for further 180 epochs to ensure convergence (see Appendix for details). 
Our experimental setup is schematized in Fig~\ref{fig:blur_deficit} (top panel).
Here we focus on the simple CIFAR-10 classification dataset, and we later examine different architectures and datasets, and learning approaches.

At the end of the training, both sensors are working well and contain partially disjoint information about the task variable, so the network would benefit from using both of them.
However, in Fig.~\ref{fig:blur_deficit} we see from the RSV that weakening the right sensor by blurring it during the initial transient will permanently inhibit its use even after removing the deficit. 
More specifically, at the end of normal training units in the network attend equally to either sensor (leftmost panel). However, in the network trained with a short deficit the neurons only encode information about the ``initially good'' left sensor (the RSV of the units concentrates around -1, rightmost panel). This mirrors the occular dominance findings present in monkeys with a cataract \cite[Fig.~7]{wiesel1982postnatal}. Similarly, the longer the deficit is present during the initial training, the more the downstream performance on the CIFAR-10 classification task is impaired (Fig.~\ref{fig:depth}, left). However, the reduction of performance is not as drastic as the RSV change, since the network can compensate and achieve a good accuracy on the task using only the good sensor.

\textbf{Dependency on depth.} In Sect.~\ref{sec:theory} we note that depth is fundamental to make critical periods emerge in multi-sensor networks. We further claim that increasing the depth of the network makes critical periods more evident. Indeed, in Fig.~\ref{fig:depth} (center) we show that increasingly deeper network have increasingly more marked permanent impairment as a result of a temporary deficit.

\subsection{Learning synergistic information}

We have seen that temporary weakening of one sensor may completely inhibit its learning. We now consider an alternative deficit where the two sensors are both working well, but are initially trained on uncorrelated data and only later trained together. This situation is common in every day machine learning, for example when pre-training backbones on different modalities separately (e.g., a text and a vision backbone) and then fine-tuning them together on a downstream task.

\textbf{Dissociation deficit.} To keep the two modalities symmetrical, we consider a similar set up as before where we feed to each pathway of a network the left and a right crop of an image. Both crops are now always full-resolution. However, we introduce a \textit{dissociation} deficit, during which the right crop is sampled from a different image than the left one. During the dissociation, the task is to predict either the class of the left image or the right image with probability $0.5$. This deficit removes any synergistic information between the two pathways, but still encourages the two pathways to extract any unique information from the inputs. 

We observe that this setup too has a critical period: In Fig.~\ref{fig:strabismus},  we see that, after normal training, the units are equally sensitive to both the left and right inputs  (histogram clusters around zero). However, after training with an increasingly longer dissociation deficit, the histogram becomes increasingly polarized around $\pm 1$, suggesting that each unit is encoding information only about the right or the left image. This precludes the possibility that the network is extracting synergistic information from the two views (which would entail units that process information from both sensors).  This mirrors the ocular dominance representations observed in strabismic monkeys  \cite[Fig.~10-12]{wiesel1982postnatal}. Similarly to the dissociation deficit, in strabismus, the eyes are not aligned, thus breaking the normal correlation between the views. The dissociation deficit also produces a permanent impairment in the downstream performance (Fig.~\ref{fig:depth}, right) but again the effect is not as drastic as in the RSV plot since the network compensates by using each pathway separately (albeit synergistic information is lost).

\begin{figure*}[t]
    \centering
    \includegraphics[width=0.32\textwidth]{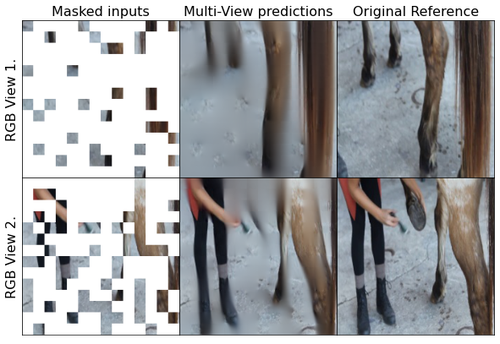}
    \includegraphics[width=0.32\textwidth]{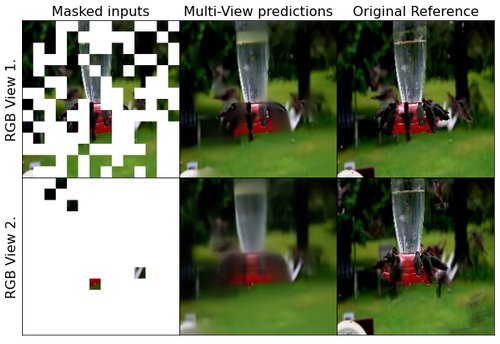}
    \includegraphics[width=0.32\textwidth]{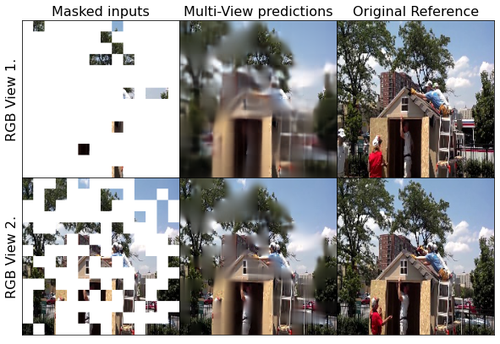}
    \caption{ \textbf{Top.} Example inputs (left column), reconstructions (middle columns), and original targets (right columns) for the Multi-View Transformer, with random sampling of patches from the two views. Note that the model can reconstruct missing information from one view using the other.
    \vspace{-1em}
    }
    \label{fig:multi-view-transformer}
\end{figure*}

\subsection{Synergistic information in videos}
\label{sec:video-supervised}

So far we have seen that supervised deep networks, similar to humans and animals, have critical periods for learning correspondences between multi-view data. We confirmed this both at the behavioural (measured in terms of performance and visual acuity for the deep networks and animals respectively) and at the representation level, quantified by the neuron sensitivity. We now  investigate whether such phenomenon generalize across learning strategies, architecture, and tasks. 

\begin{figure}
    \centering
    \includegraphics[width=0.33\textwidth]{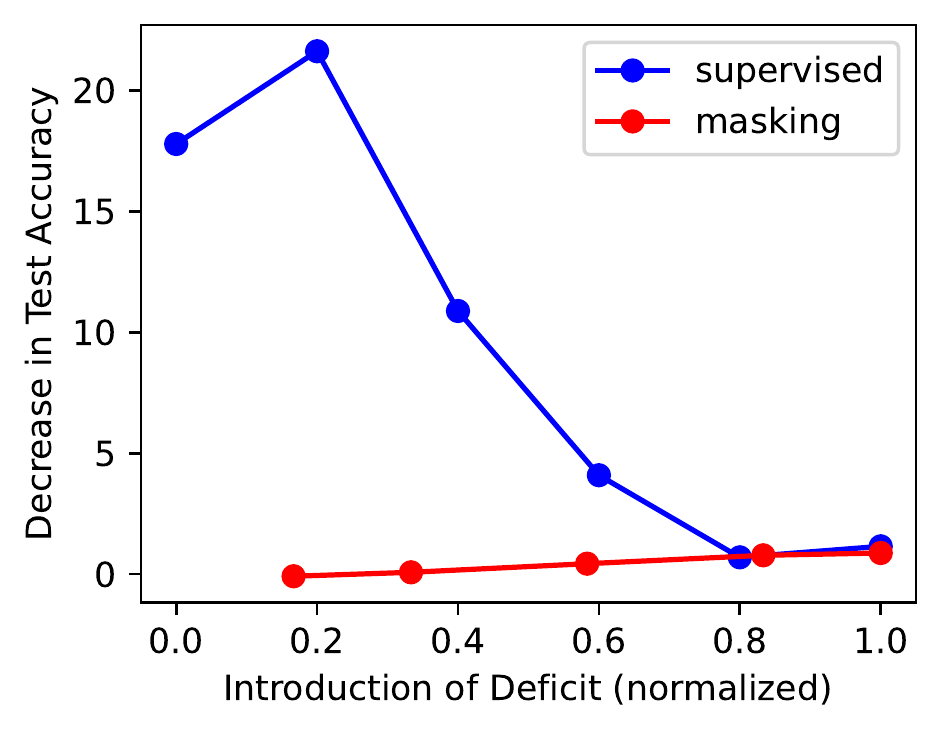}
    \hspace{2em}
    \caption{\textbf{Masking objective with cross-sensor reconstruction loss does not exhibit a critical learning period.} We found that the unsupervised network was much more robust to perturbations early in the training (red trace), whereas that supervised objective was not (blue trace). 
 \vspace{-1em}
 }
   \label{fig:sensitivity}
\end{figure}

\textbf{Multi-View Transformer.} Aside from integrating information from different sensors, animals and artificial networks need to be able to integrate information through time. We can think of frames of a video as being different views or sources of information that are correlated through time, and we can study how a network learns to integrate such information. We opted to use a more flexible transformer-based visual architecture, which has recently achieved state-of-the-art results in computer vision tasks \cite{dosovitskiy2020image, he2021masked}, and language tasks \cite{vaswani2017attention, devlin2018bert}. Visual transformers  are typically trained either with a supervised loss \cite{dosovitskiy2020image} or a masking-based objective, followed by fine-tuning \cite{he2021masked}. We focus now on the first case, and analyze the second in the next section.
In order to process multiple frames of a video, we use a modified Multi-Modal Masked Auto-Encoder \cite{bachmann2022multimae}, which we train in a fully supervised fashion. We refer to this as a \emph{Multi-View Transformer}. %

To capture multiple views of a scene, we opted to use the the Kinetics Action classification video dataset \cite{carreira2017quo}, which consist in classifying one of 400 possible actions given a video clip. To adapt the task to our setting, from each video we select two random frames that are a multiple of $0.33$ seconds apart to comprise our two views, and feed them to the \emph{Multi-View Transformer}. Due to their temporal correlation, the two frames together contain more information (the motion) than either frame individually. We use a similar dissociation deficit as in the previous section: During the dissociation deficit period, we sampled the two frames from independent videos in order to break their temporal correlation. In this case, the classification label coming from either view with $p = 0.5$ (see Appendix for training details). We introduce the deficit in a sliding window of fixed size starting, and vary the starting time to measure the sensitivity of each part of the training process.

Even on a largely different architectures (transformer instead of ResNet) and a more complex task (action classification on natural video instead of CIFAR-10), in Fig.~\ref{fig:sensitivity} we observe the same trends as in the previous section. 
Training with a temporary dissociation deficit permanently prevents the network from extracting synergistic temporal information from the frames. 
Unlike in the previous experiment, since the synergistic information is fundamental for the action classification task, the network cannot compensate the deficit and perturbations during the critical period also results in an harsh decrease of up to 20\% in the final test accuracy (Fig.~\ref{fig:sensitivity}, blue trace).

\subsection{Overcoming critical periods with cross-sensor reconstruction}
\label{sec:video-unsupervised}

Our previous experiments suggest that critical periods can be caused by competition between sensors which increases the selectivity of the units. If this is the case, we may hypothesize that training adding a cross-sensor reconstruction objective may help forcing the unit to learn how to encode cross-sensor information.
To test this hypothesis, we train the Multi-View transformer of Sec.~\ref{sec:video-supervised} using the cross-sensor masking-based reconstruction objective of \cite{bachmann2022multimae} and compare it with the supervised case. 
The self-supervised masked-image reconstruction task could encourage correspondences to be learned (if un-occluded parts of one view are helpful for reconstructing the other view), and may force learning synergistic information irrespective of the initial transient.
In Fig.~\ref{fig:multi-view-transformer}, we show that indeed the masking-based pre-training is successful in using information from one source to predict masked patches of the other.

We train using the same protocol as Sec.~\ref{sec:video-supervised} to pre-train the Multi-View Transformer using the cross-reconstruction objective. We then subsequently fine-tuned for $20$ epochs on the downstream supervised classification task (see Appendix for details). %
In Fig.~\ref{fig:sensitivity} we see that the unsupervised network was much more robust to perturbations early in the training, whereas that supervised objective was not. 
To understand whether such robustness was due to large changes to the representation when fine-tuning, we applied the RSV on the output of the encoder's representation and found that while the resulting distribution became slightly more symmetrically balanced, it retained a similar bimodal distribution to the pre-trained representation. (Fig.~\ref{fig:sensitivity_rsv}).

\vspace{-0.2em}

\section{Discussion}

We have shown -- in a variety of architectures and tasks -- the existence of critical learning periods for multi-source integration: a temporary initial perturbation of an input source may permanently inhibit that source, or prevent the model from learning how to combine multiple sources. These trends replicate similar phenomena in animals, and point to the underlying complexity and brittleness of the learning dynamics that allow a network (or an animal) to fuse information. To simplify the analysis of the learning dynamics, we focused  on tasks with homogeneous sources (stereo, video). We leave to future work to further study the role played by the asymmetry between sources (e.g., different informativeness or ease).
Our theoretical and empirical analysis leads to several suggestions: Pre-training different backbones separately on each modality, as advocated in some foundational models, may yield representations that ultimately fail to encode synergistic information. Instead, training should be performed across modalities at the outset. On the theoretical side, our work suggests that  analysis ``at convergence'' of the learning dynamics of a network are irrelevant for sensor fusion, as their fate is sealed during the initial transient learning. It also suggests that conclusions drawn from wide and shallow networks may not transfer to deep networks in current use.

{\small
\bibliographystyle{CVPR-Papers/ieee_fullname}
\bibliography{bibliography}
}

\clearpage
\newpage
\appendix
\onecolumn
\section{Supplementary Material}

\noindent Our code is available at: \href{https://github.com/mjkleinman/CriticalPeriodMultiView}{https://github.com/mjkleinman/CriticalPeriodMultiView}.

\subsection{Description of simulated RSV distributions}
\label{sec:rsv_simulation}
When evaluating the RSV on a synthetic distribution, we considered the following generative model that consists of a common component $x_0$ with additive noise:
\begin{equation}
\begin{gathered}
    x_a = x_0 + n_a, \quad x_b = x_0 + n_b,\\
    z_i = w_i x_a + (1 - w_i) x_b, \\
    w_i \sim \text{Beta}(\alpha, \beta) ,\quad x_0 \sim \mathcal{N}(0, ~ 1), \quad n_a \sim \mathcal{N}(0, ~ 1),  \quad n_b \sim \mathcal{N}(0, ~ 1).\
\end{gathered}  
\end{equation}
Depending on the values of $\alpha$ and $\beta$, the Beta distribution that the weights $w_i$ are drawn from will take different shapes, changing how units in the representation $z$ vary with inputs $x_a$ and $x_b$. We find that the distribution of RSVs in Fig.~\ref{fig:fsv_schem} reflect the full spectrum of these various distributions, where the resulting RSVs can vary from an approximately Gaussian distribution where units vary equally with both modalities, to polarized representations where units vary uniquely with one modality

For this synthetic simulation, we can derive a closed form expression for the RSV. In particular (and dropping the subscript $i$ for clarity), 
\begin{equation}
    z = x_0 + w n_a + (1 - w) n_b
\end{equation}
and note that $z$ will be distributed as a normal distribution. Then,
\begin{align}
    SV_i &= Var(Z|X_a = x_a) \\
    &= \sigma_z^2 (1 - p^2) \\
    &= \sigma_z^2 (1 - \frac{Cov(z,x_a)^2} {\sigma_z^2 \sigma_{x_a}^2})    
\end{align}
We know that 
\begin{equation}
    \sigma_z^2 = \sigma_{x_0}^2 + w^2\sigma_{a}^2 + (1-w)^2 \sigma_b^2
\end{equation}
since $x_0$, $n_a$, and $n_b$ are independent. Finally, 
\begin{align}
    Cov(z, x_a) &= \E[(Z - \E[Z])(X_a - \E[X_a]] \\
    &= \E[ZX_a] \\
    &= \E[(wX_a + (1-w)X_b ) X_a] \\
    &= \E[(w(X_0 + N_a) + (1-w)(X_0 + N_b)) (X_0 + N_a)] \\
    &= \E[(X_0 + wN_a + (1-w)N_b) (X_0 + N_a)] \\
    &= \E[X_0^2] + w \E[N_a^2] \\
    &= \sigma_{x_0}^2 + w\sigma_a^2
\end{align}
We also know that
\begin{equation}
     \sigma_{x_a}^2 = \sigma_{x_0}^2 + \sigma_a^2.
\end{equation}
We can then solve for $SV_i$ by plugging Eq 9, 16, 17 into Eq 8 and obtain:
\begin{align}
SV_i &= \sigma_z^2 (1 - \frac{Cov(z,x_a)^2} {\sigma_z^2 \sigma_{x_a}^2}) \\
 &= (\sigma_{x_0}^2 + w^2\sigma_{a}^2 + (1-w)^2 \sigma_b^2) (1 - \frac{\sigma_{x_0}^2 + w\sigma_a^2}{(\sigma_{x_0}^2 + \sigma_a^2) (\sigma_{x_0}^2 + w^2\sigma_{a}^2 + (1-w)^2 \sigma_b^2)})
\end{align}

We assumed that the representation $z_i$ for half of the units were sampled from above generative model, while the other half the representation $z_i$ were sampled from the reverse convex combination of inputs, i.e, $z_i = w_i x_b + (1-w_i) x_a$.

For simulations 2-4, we set $\beta=20$ and varied $\alpha$ in $[1, 20, 30]$ respectively. We considered a representation on $N = 20 000$ units. For the first simulation we only considered the half of units in the generative model above, with $\alpha=1$ and $\beta=10$.

\subsection{Generalization of RSV to arbitrary number of sensors}

We can naturally generalize the RSV to an arbitrary number $n$ of sources. To do so, define:
\begin{equation*}
    SV_i(X_j, x_1,...,x_{j-1},x_{j+1},...,x_n) = Var(f(\mathbf{X})_i|X_1=x_1, ..., X_{j-1}=x_{j-1}, X_{j+1}=x_{j+1},..., X_n = x_n),
\end{equation*}

and then collect the individual source variances into a vector $\mathbf{SV}_i$ of size $n$. Then normalized sensor variance would be
\begin{equation*}
    RSV_i = \text{softmax}(\mathbf{SV_i}),
\end{equation*}
which provides a normalized quantification (between $0$ and $1$) of how much an individual unit varies with each sensor modality $j$.

\subsection{Description of deep linear network experiment}
We considered the original input-output correlation (before dropping a sensor) to be
\begin{equation}
    \sio_{pre} = 
\begin{bmatrix}
1 & 0 & 3 & 1 & 0 & 0 & 0 & 0 \\
1 & 0 & 0 & 0 & 1 & 0 & 0 & 0 \\
0 & 1 & 0 & 0 & 0 & 1 & 0 & 0 \\
0 & 1 & 0 & 0 & 0 & 0 & 1 & 0 \\
0 & 0 & 3 & 0 & 0 & 0 & 0 & 1
\end{bmatrix}
\end{equation}
Our perturbation involved dropping a sensor, in this case the third column, leading to
\begin{equation}
    \sio_{post} = 
\begin{bmatrix}
1 & 0 & 0 & 1 & 0 & 0 & 0 & 0 \\
1 & 0 & 0 & 0 & 1 & 0 & 0 & 0 \\
0 & 1 & 0 & 0 & 0 & 1 & 0 & 0 \\
0 & 1 & 0 & 0 & 0 & 0 & 1 & 0 \\
0 & 0 & 0 & 0 & 0 & 0 & 0 & 1
\end{bmatrix}
\end{equation}

Using the analytical equations for the learning dynamics given by \cite{saxe2019mathematical} for the shallow and deep network, we investigated how learning the task (row 5) was affected (Fig.~\ref{fig:deep_linear}), finding that such a perturbation had a significant on the dynamics of sensor learning in the deep, but not shallow, network.

\subsection{Description of architectures and training}

Most of our experiments are based on the ResNet-18 architecture \cite{he2015deep}. We modified the architecture to process multi-sensor input with what we call a SResNet-18. We separately process two  initial pathways which we combine in an additive manner. In particular, the initial pathway followed the architecture of \cite{he2015deep} directly up to (and including) $\texttt{conv3\_x}$ (See Table 1 of \cite{he2015deep}). After combining the pathways, the remaining layers followed the ResNet-18 architecture directly. 

To examine the effect of depth, we modified the All-CNN architecture \cite{springenberg2014striving}, following \cite{achille2018critical}. In particular we processed each pathway with the following architecture:

\begin{center}
\texttt{conv $96$ - [conv $96 \cdot 2^{i-1}$ - conv $96 \cdot 2^i$ s2]${}_{i=1}^n$ - conv $96\cdot 2^n$ - conv1 $96\cdot 2^n$ - conv1 $10$
}
\end{center}
where $s$ refers to the stride. We then merged the final representation from each pathway in an additive manner. We examined the setting when $n = 1,2,3$. We used a fixed learning rate of $0.001$ in these experiments.

\subsection{Description of Blurring Experiments (Fig.~\ref{fig:blur_deficit})}

We attempted to simulate a cataract-like deficit by blurring the image to one pathway. We reduced the resolution of the image being passed to one pathway by first resizing the Cifar images to $8 \times 8$, and then resizing to its original size ($32 \times 32$ pixels, decreasing the available information.

While training, we applied standard data augmentation on the uncorrupted pathway (random translation of up to $4$ pixels, and random horizontal flipping. We then retained a width $w$ of the leftmost and rightmost pixels from uncorrupted and corrupted pathway respectively, setting $w = 16$ unless otherwise stated. At inference time, no data augmentation was applied and the leftmost $w$ pixels and rightmost $w$ pixels was supplied to each pathway respectively. We used an initial learning rate of $0.075$, decaying smoothly at each epoch with a scale factor of $0.97$. We also found that using a fixed learning rate of $0.0005$ (Fig.~\ref{fig:fixed-lr}) and different initial learning rates (Fig.~\ref{fig:robustness}, right) had similar RSV and performance changes as a result of the initial deficits.

To quantify the information contained in the representation, we randomly masked out each pathway with $p = 0.1$ during training, and computed the usable information $I_u$ contained in the representation $Z$ abbout the task $Y$ following \cite{kleinman2020usable, Xu2020theory} by computing $I_u(Z;Y) = H(Y) - L_{CE}$, with $H(Y)$ being known and equal to $\log_2 10$ since the distribution of targets is uniform, and $L_{CE}$ being the cross-entropy loss on the test set. We reported the corresponding RSV plots, and network performance in Appendix Fig.~\ref{fig:blur_deficit_info}, which reveal similar performance trends and polarization of units, when pre-training with the random masking as in Fig.~\ref{fig:blur_deficit}.

\subsection{Description of Independent Pathways Experiment (Fig.~\ref{fig:strabismus})}

We followed the same setup as above, but instead randomly permuted the images fed to the `right' pathway across the batch, breaking the correlation between the views. We trained using an initial learning rate of $0.05$, decaying smoothly with a scale factor of $0.97$. When training with the deficit we randomly sampled the target from the different views with $p=0.5$. We also modified the architecture to produce multiple classification outputs, corresponding to a classification based on both views, or each pathway respectively. This modification was helpful for interpreting the polarization plots. While training, the loss function was applied on the head that contained the proper input-target correspondence. After the deficit, and during inference, only the head corresponding to both views was used.

\subsection{Description of Masking + Supervised MultiViT training}

These experiments were based on the MultiMAE architechture \cite{bachmann2022multimae}, using their implementation and closely following their default settings. We adapted their implementation to process two separate RGB views coming from Kinetics-400 dataset \cite{carreira2017quo}. We used a patch size of $16$ in all experiments, and the AdamW optimizer \cite{loshchilov2018decoupled}. All inputs were first resized to $224 \times 224$ pixels. Our learning rate followed the linear scaling rule \cite{goyal2017accurate}.

For the masking sensitivity experiments in Fig.~\ref{fig:sensitivity}, we used a fixed delay of $1.33$ seconds ($4$ frames) between frames, and trained with an initial base learning rate of $0.0001$, with $40$ epochs of warmup for the learning rate. 
We trained for $800$ epochs, with a $200$ epoch deficit of independent frames during the pre-training starting at different epochs during training. We used a masking ratio of $0.75$. We pre-trained with a batch size of $256$ per GPU on $8$ GPUs. After the pre-training, we fine-tuned for $20$ epochs with all the tokens and the corresponding action classification label. We fine-tuned on $8$ GPUs with a batch size of $32$. We fine-tuned with a learning rate of $0.0005$, with $5$ epochs of warmup.

For the supervised experiments, we trained our networks with an initial base learning rate of $0.01$ for $120$ epochs using all the tokens, with $20$ epochs of warmup. We applied a temporary deficit of independent frames for $20$ epochs, starting at various epochs during the training. We used in cutmix ($1.0$) and mixup ($0.8$) applied to each view) while training and we used a random baseline between frames. For the supervised experiments, we used a batch size of $64$ per GPU.

In both the masking and supervised experiments in Fig.~\ref{fig:sensitivity}, we reported the difference of networks trained with a deficit starting at different epochs of training against a corresponding model trained without any deficit. In Fig.~\ref{fig:multi-view-transformer}, we show example reconstructions from our Multi-View transformer pre-trained without a deficit for $800$ epochs with a random baseline between frames.

\section{Additional Plots}
\newpage

\begin{figure}[t!]
    \centering
    \includegraphics[width=0.28\textwidth]{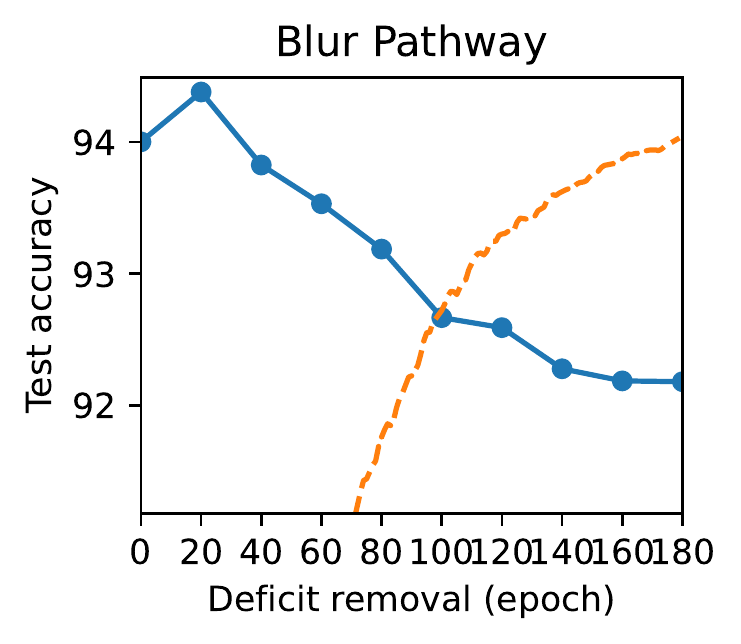}
    \\
    \includegraphics[width=0.15\textwidth]{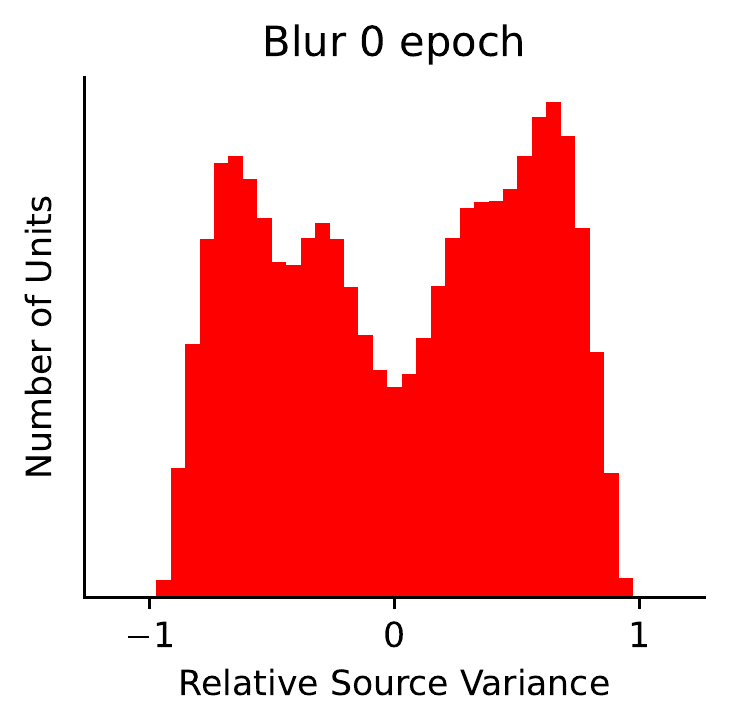}
    \includegraphics[width=0.15\textwidth]{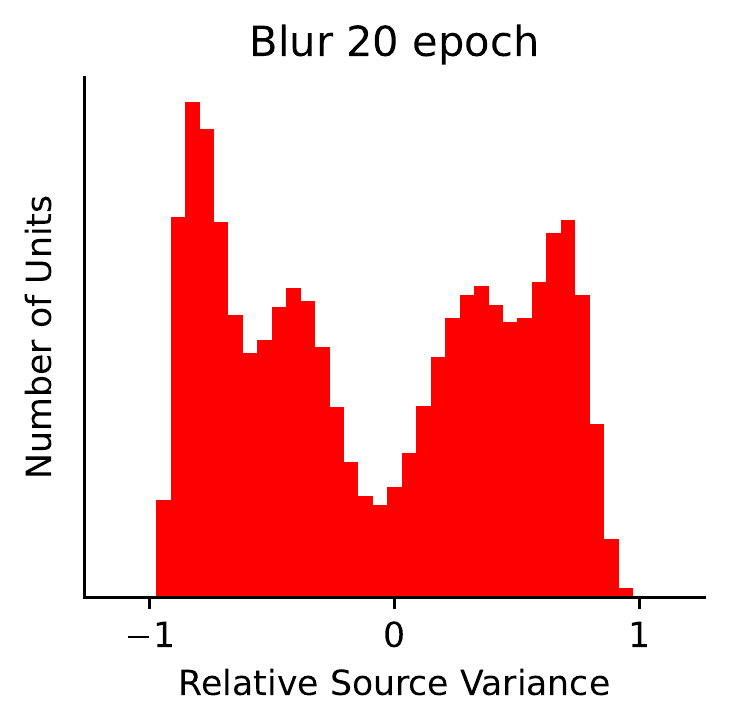}
    \includegraphics[width=0.15\textwidth]{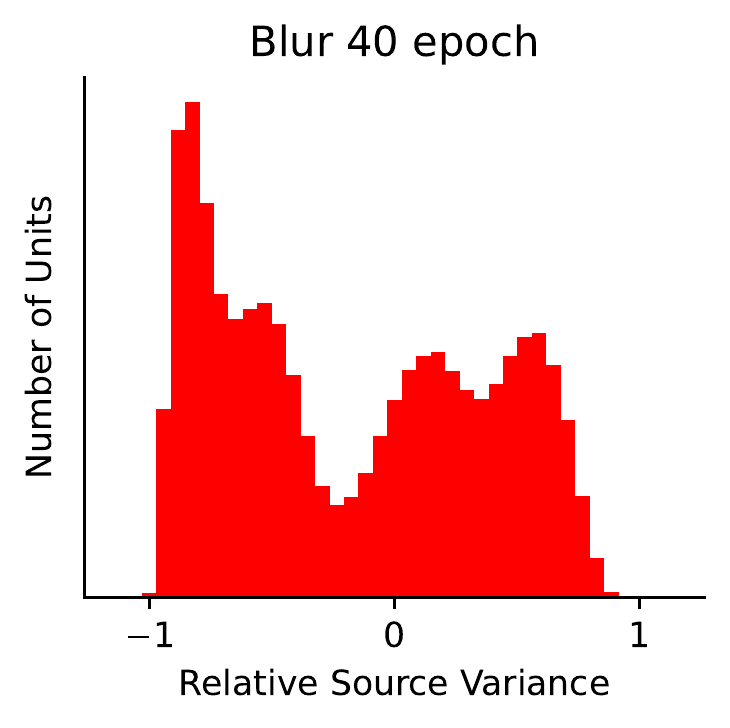}
    \includegraphics[width=0.15\textwidth]{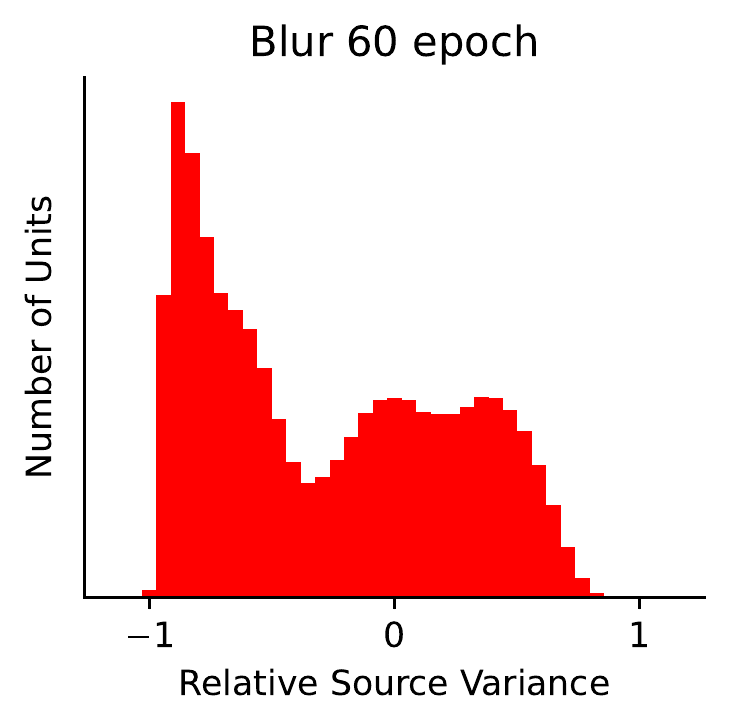}
    \includegraphics[width=0.15\textwidth]{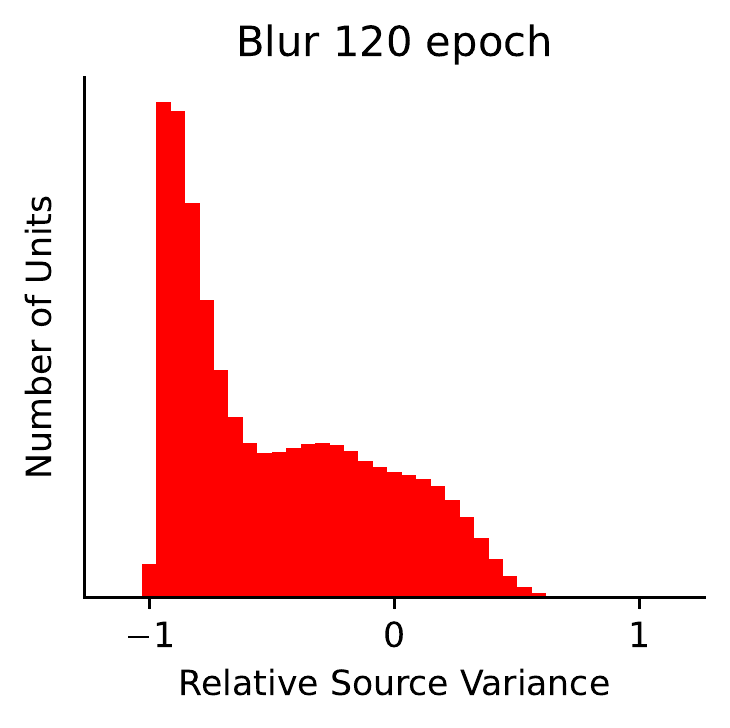}
    \includegraphics[width=0.15\textwidth]{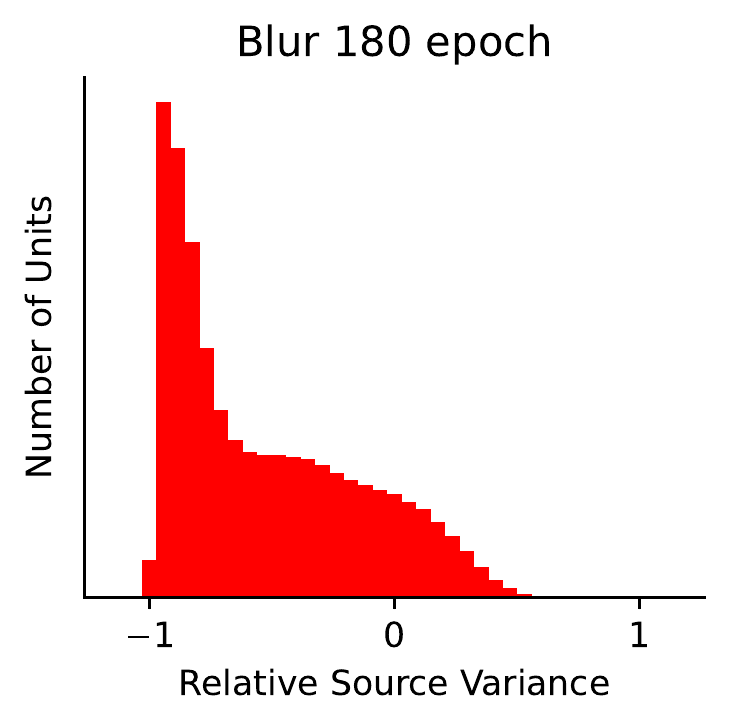}
    \\
    \includegraphics[width=0.15\textwidth]{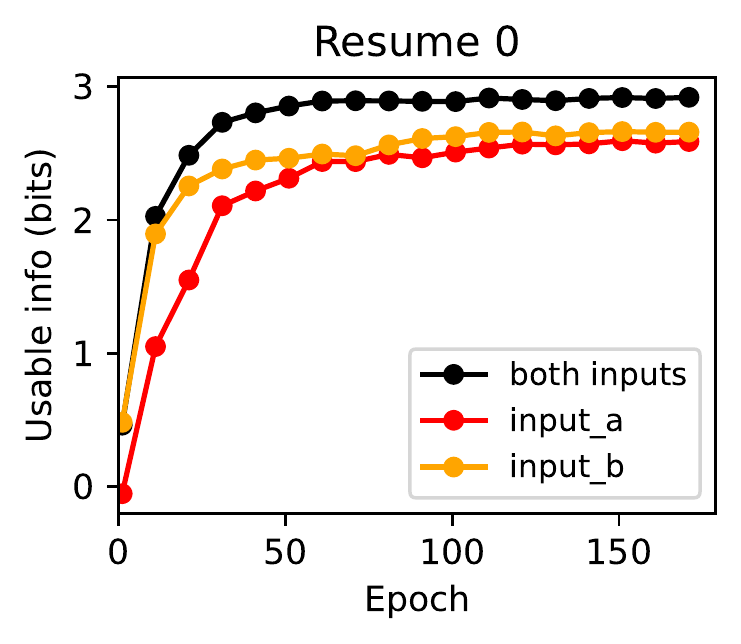}
    \includegraphics[width=0.15\textwidth]{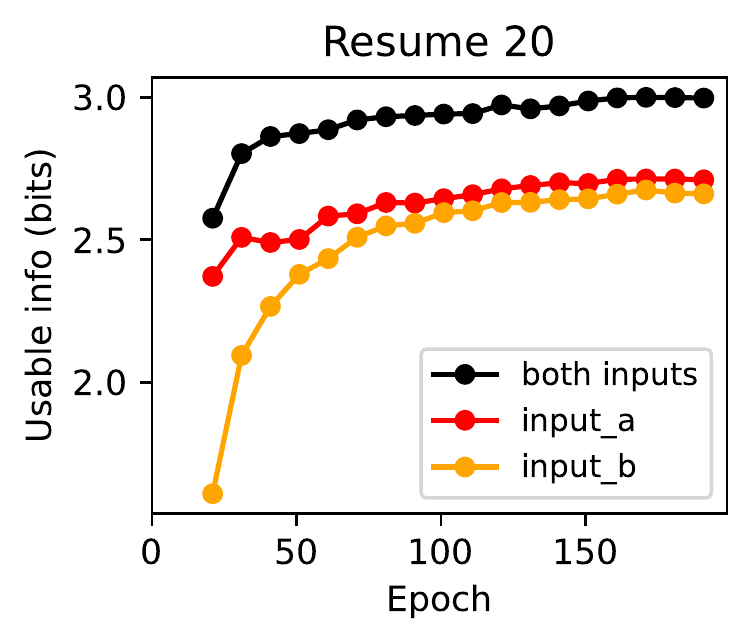}
    \includegraphics[width=0.15\textwidth]{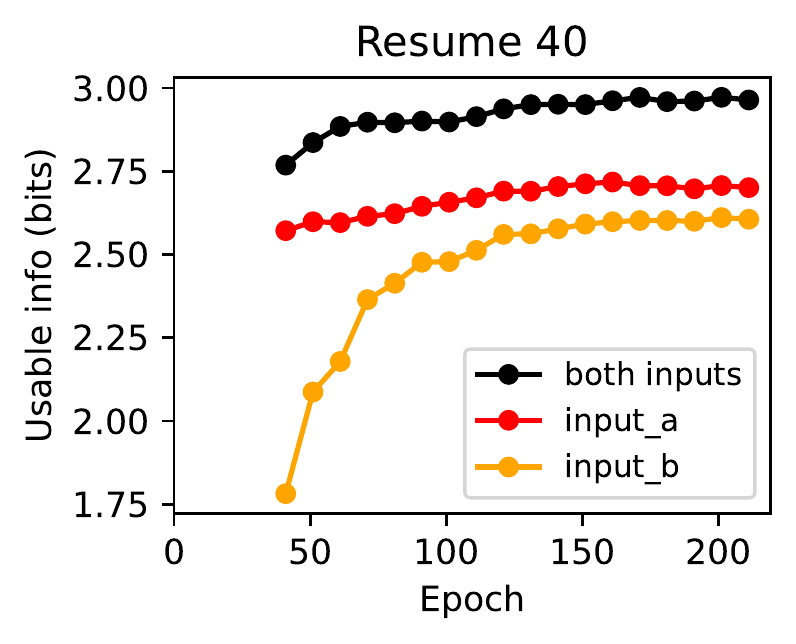}
    \includegraphics[width=0.15\textwidth]{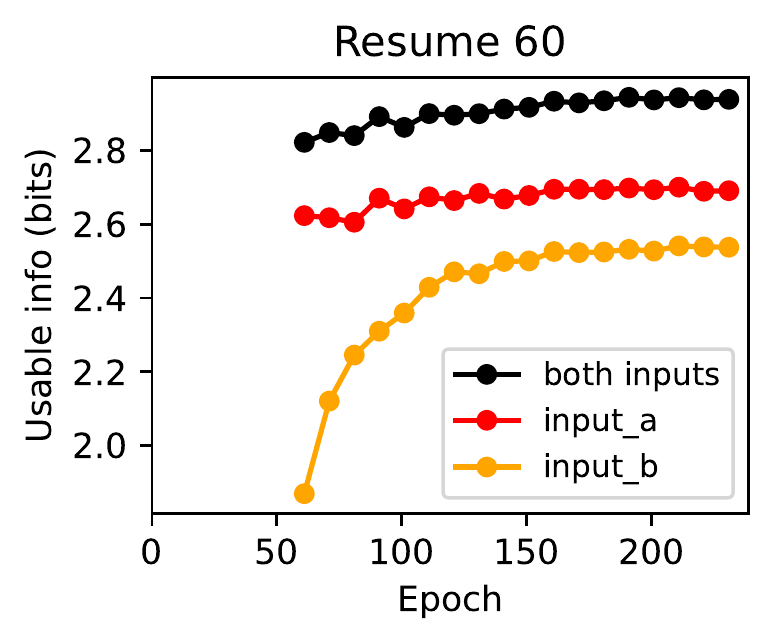}
    \includegraphics[width=0.15\textwidth]{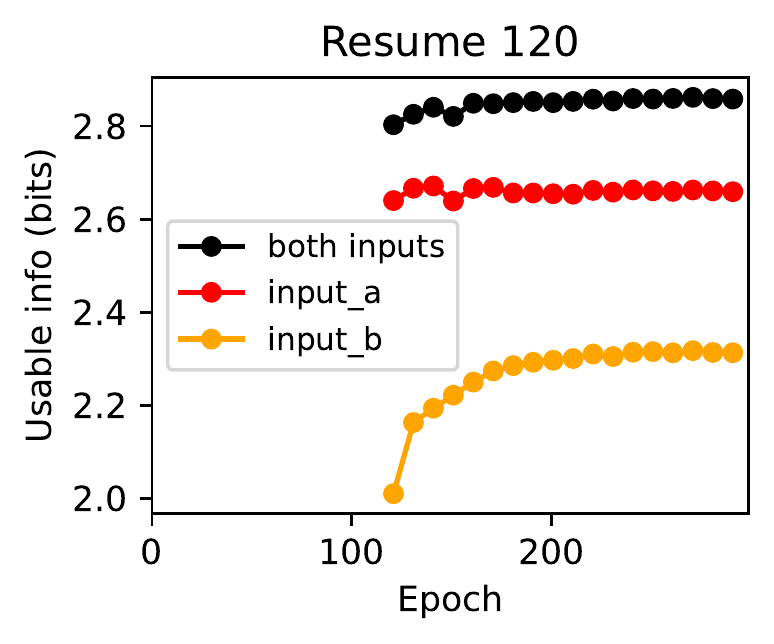}
    \includegraphics[width=0.15\textwidth]{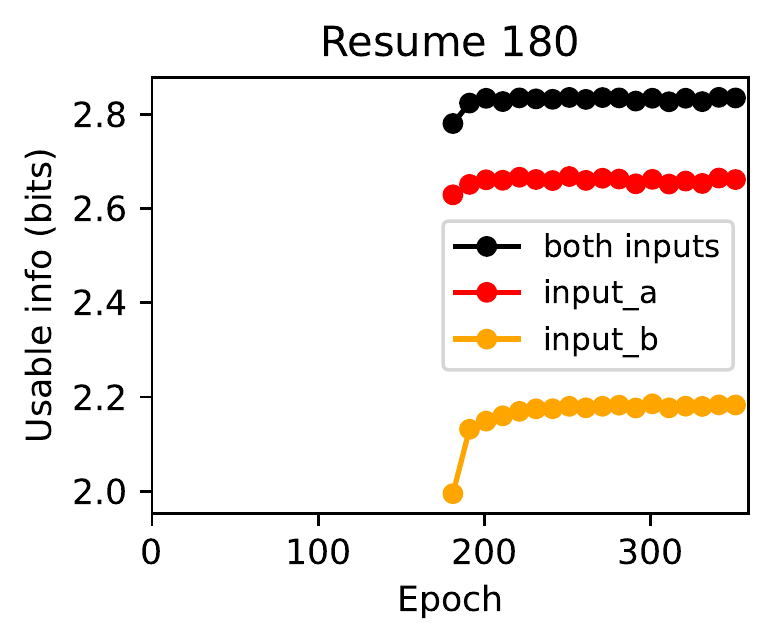}
    \caption{Same blurring experiment as Fig.~\ref{fig:depth} with corresponding Relative Source Sensitivity,  Fig.~\ref{fig:blur_deficit}, but with the addition of random masking on each view with $p=0.1$, allowing the decoding of the usable information \cite{kleinman2020usable} (bottom row). Note that the polarization (second row) is similar to Fig.~\ref{fig:blur_deficit}, which is also reflected by the inability to decode the inhibited pathway, after exposure to a sufficiently long deficit (orange trace in bottom row).
    } 
    \label{fig:blur_deficit_info}
\end{figure}

\begin{figure}[t!]
    \centering
    \includegraphics[width=0.28\textwidth]{plots_new/view16_lr0.075_sresnet_blur_diffaug_32c2bb//sresnet.pdf}
    \\
    \includegraphics[width=0.15\textwidth]{plots_new/view16_lr0.075_sresnet_blur_diffaug_32c2bb//180_resume.pdf}
    \includegraphics[width=0.15\textwidth]{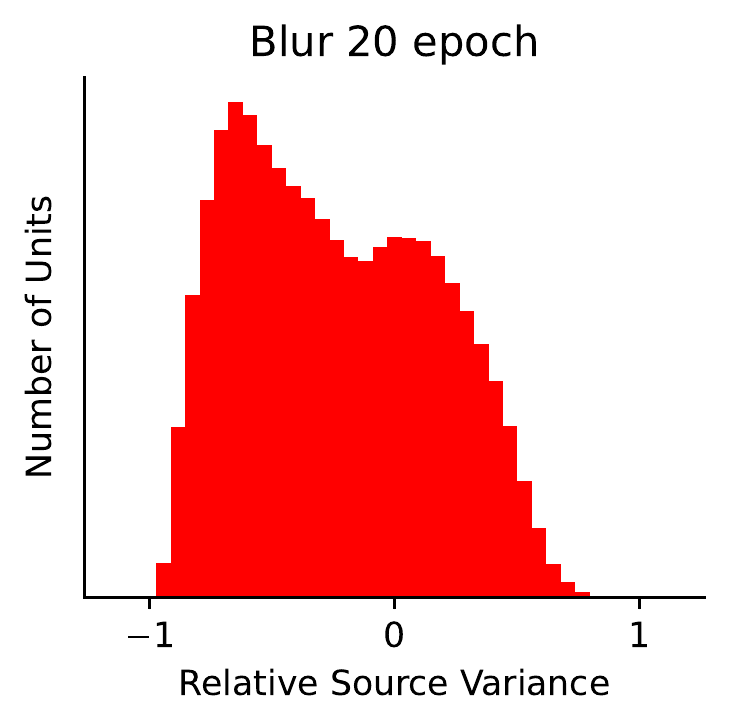}
    \includegraphics[width=0.15\textwidth]{plots_new/view16_lr0.075_sresnet_blur_diffaug_32c2bb//220_resume.pdf}
    \includegraphics[width=0.15\textwidth]{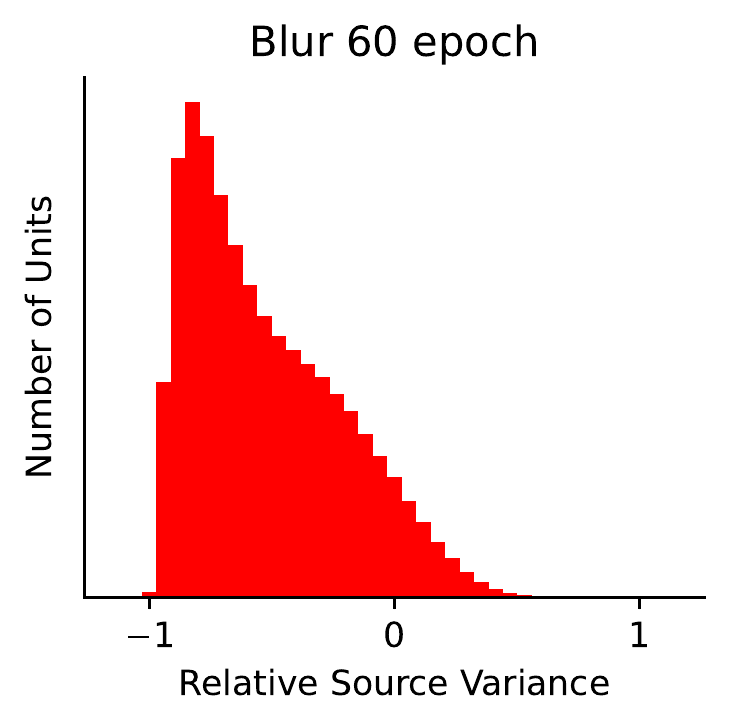}
    \includegraphics[width=0.15\textwidth]{plots_new/view16_lr0.075_sresnet_blur_diffaug_32c2bb/300_resume.pdf}
    \includegraphics[width=0.15\textwidth]{plots_new/view16_lr0.075_sresnet_blur_diffaug_32c2bb/360_resume.pdf}
    \caption{
    Same blurring experiment as Fig.~\ref{fig:depth} with corresponding Relative Source Sensitivity,  Fig.~\ref{fig:blur_deficit} for crop width of $16$ (used in the main text) for easier comparison against different crop widths in Fig.~\ref{fig:blur_deficit_14} and Fig.~\ref{fig:blur_deficit_18}.} 
    \label{fig:blur_deficit_16}
\end{figure}

\begin{figure}[t!]
    \centering
    \includegraphics[width=0.28\textwidth]{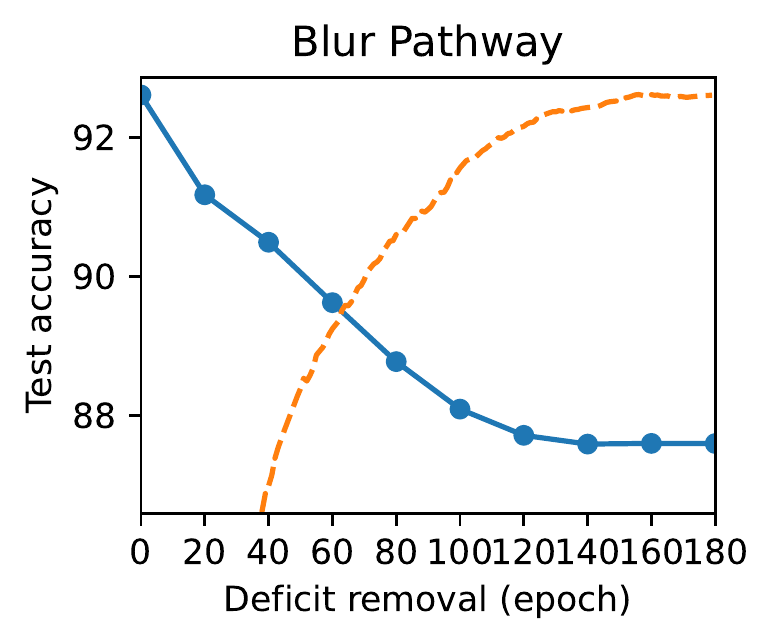}
    \\
    \includegraphics[width=0.15\textwidth]{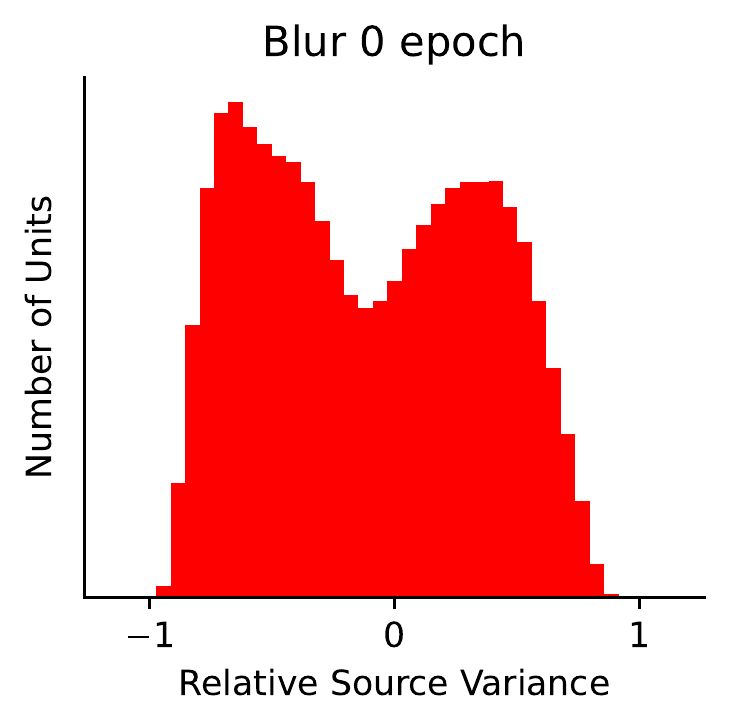}
    \includegraphics[width=0.15\textwidth]{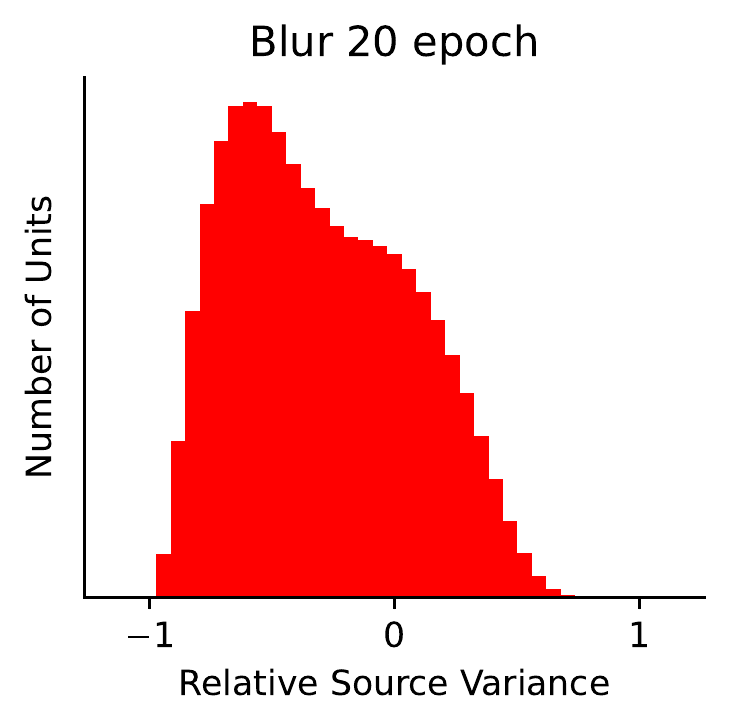}
    \includegraphics[width=0.15\textwidth]{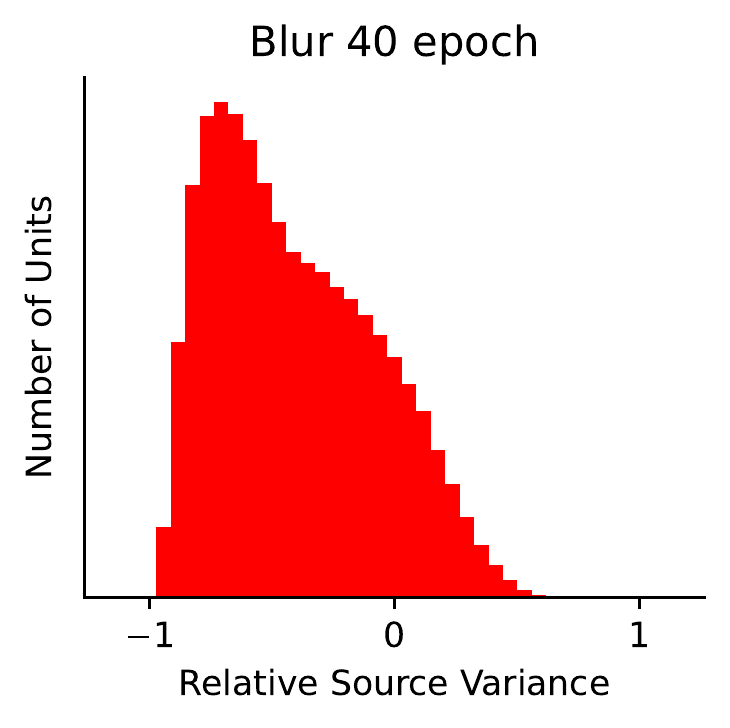}
    \includegraphics[width=0.15\textwidth]{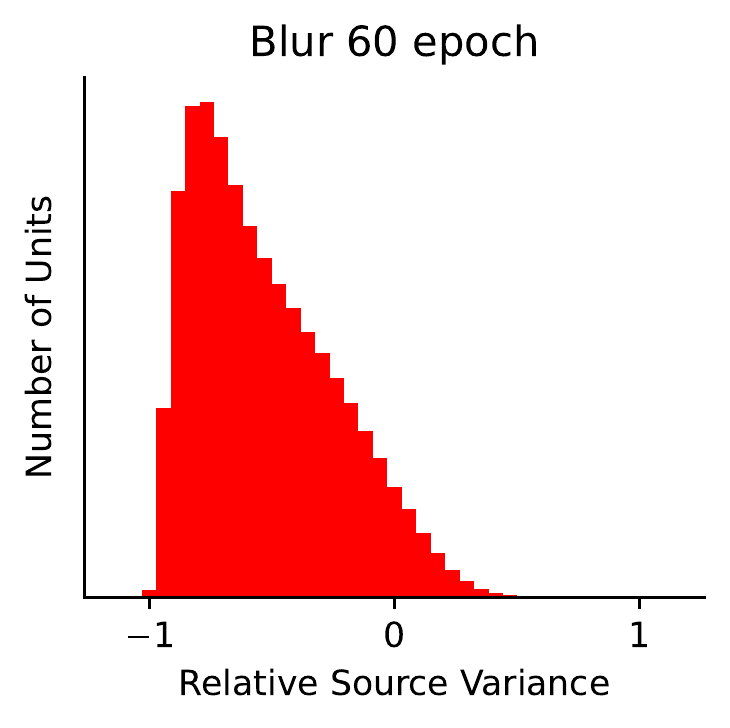}
    \includegraphics[width=0.15\textwidth]{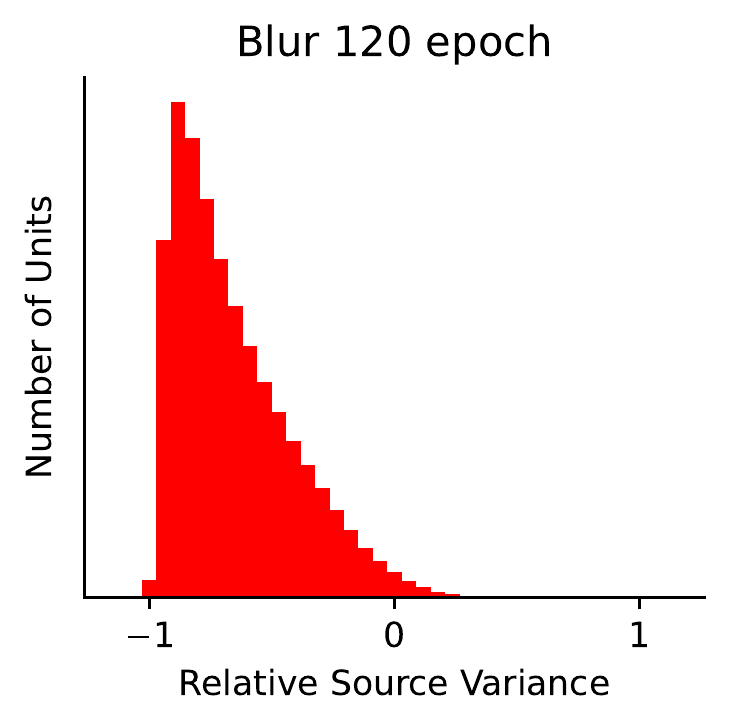}
    \includegraphics[width=0.15\textwidth]{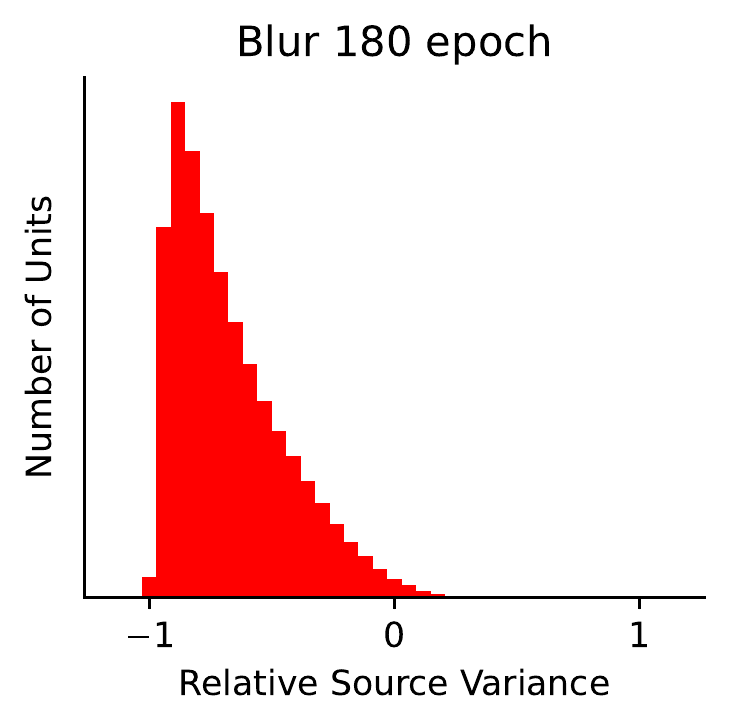}
    \caption{
        Same blurring experiment as Fig.~\ref{fig:depth} with corresponding Relative Source Sensitivity,  Fig.~\ref{fig:blur_deficit} for crop width of $14$.} 
    \label{fig:blur_deficit_14}
\end{figure}

\begin{figure}[t!]
    \centering
    \includegraphics[width=0.28\textwidth]{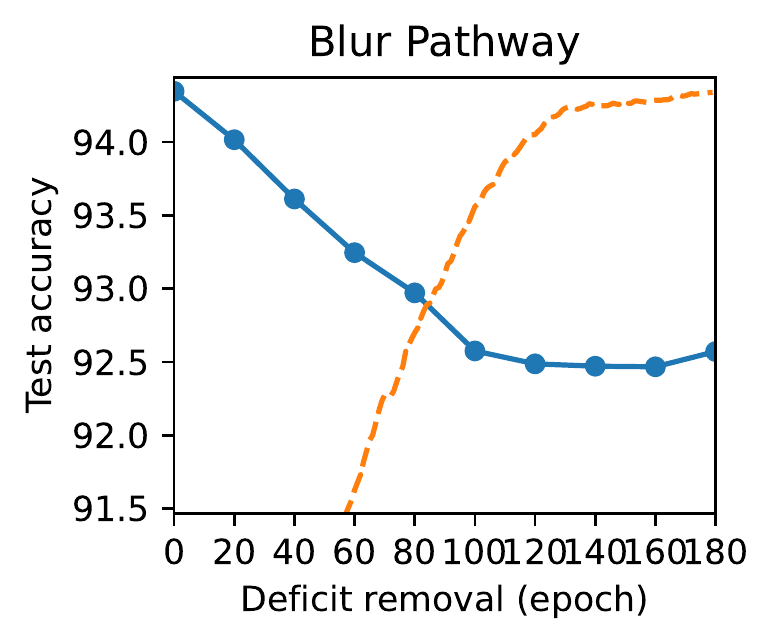}
    \\
    \includegraphics[width=0.15\textwidth]{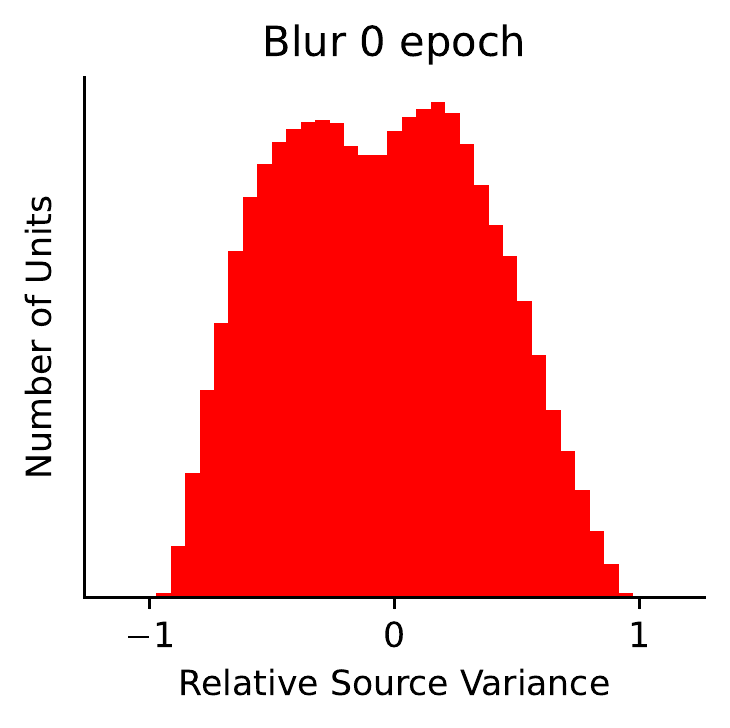}
    \includegraphics[width=0.15\textwidth]{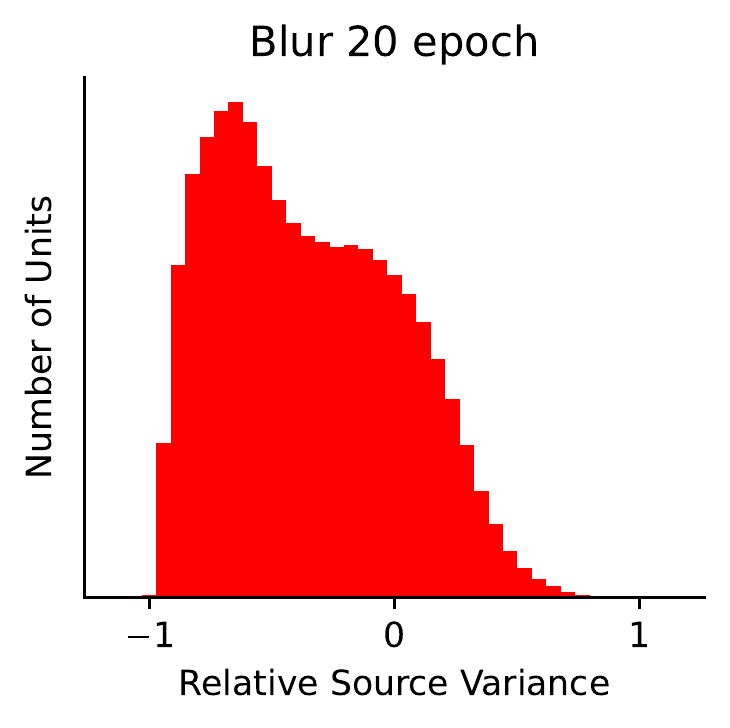}
    \includegraphics[width=0.15\textwidth]{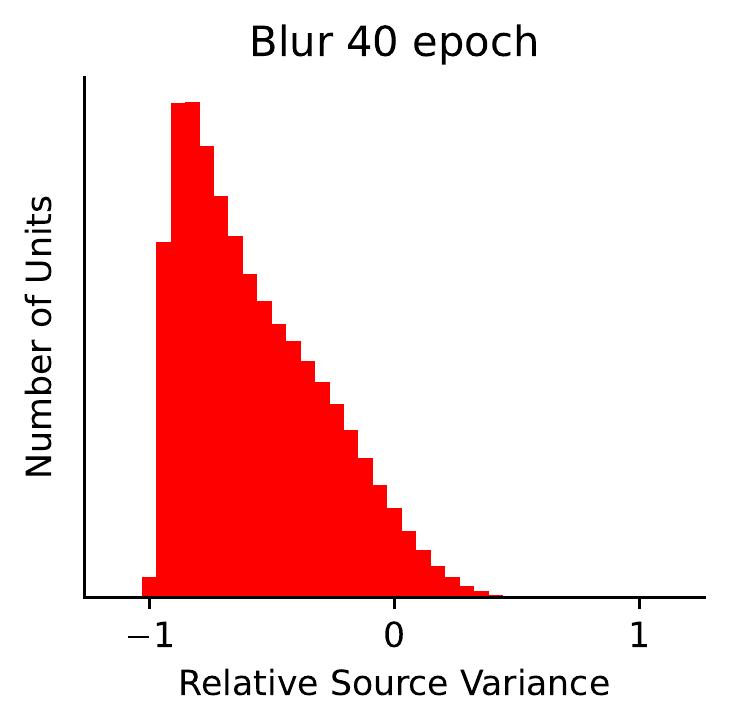}
    \includegraphics[width=0.15\textwidth]{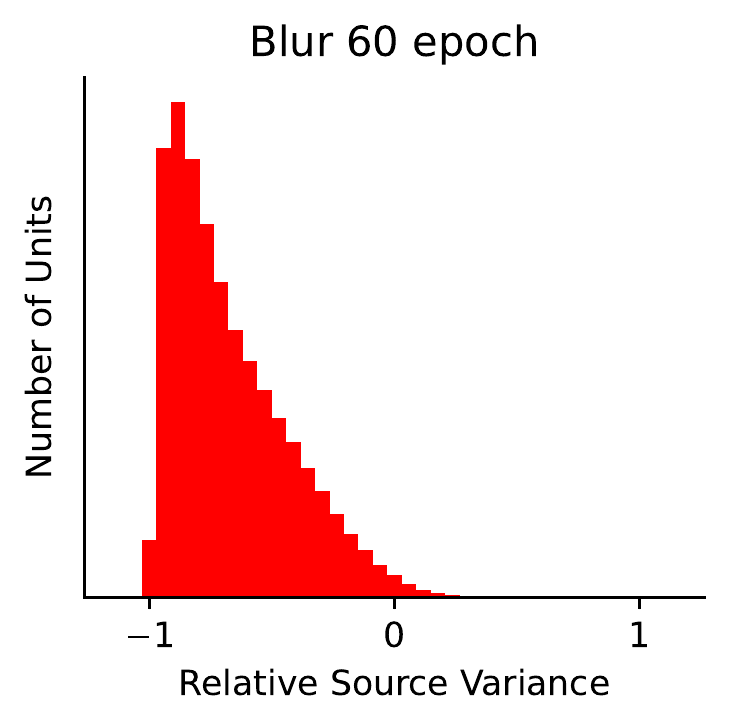}
    \includegraphics[width=0.15\textwidth]{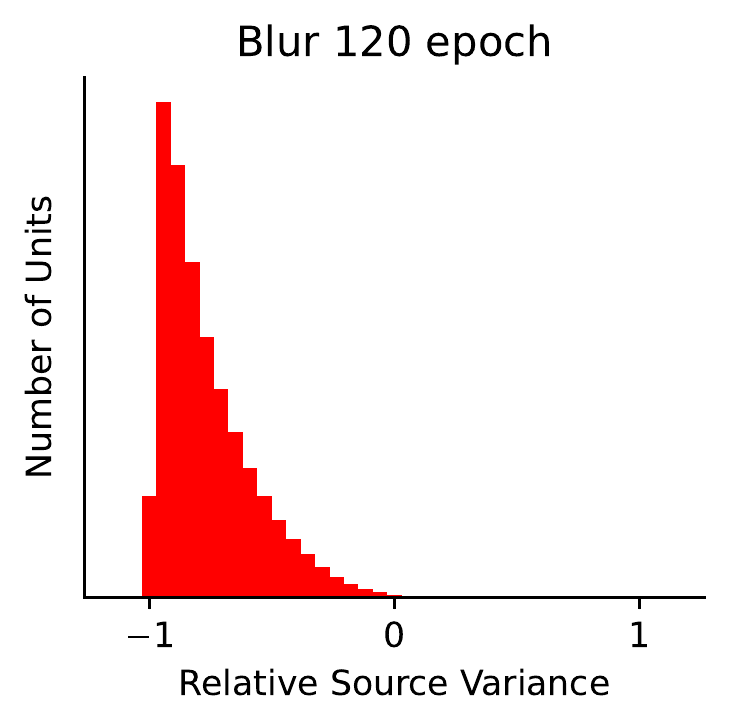}
    \includegraphics[width=0.15\textwidth]{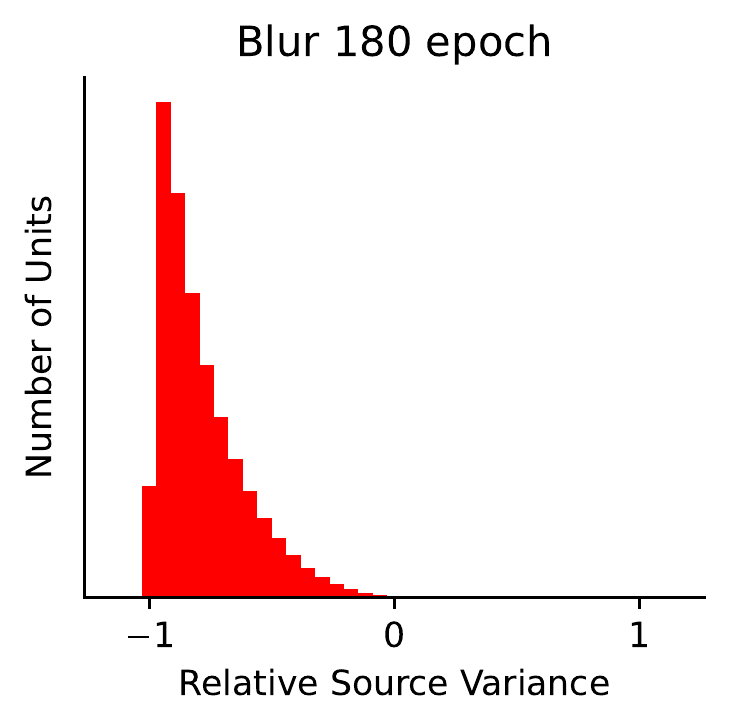}
    \caption{
        Same blurring experiment as Fig.~\ref{fig:depth} with corresponding Relative Source Sensitivity,  Fig.~\ref{fig:blur_deficit} for crop width of $18$.} 
    \label{fig:blur_deficit_18}
\end{figure}

\begin{figure}[]
    \centering
    \includegraphics[width=0.3\textwidth]{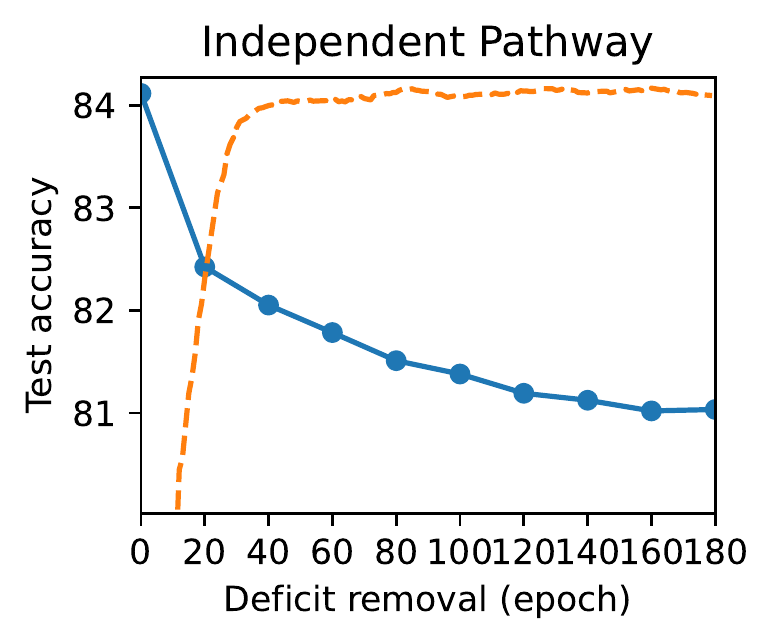}
    \\
    \includegraphics[width=0.15\textwidth]{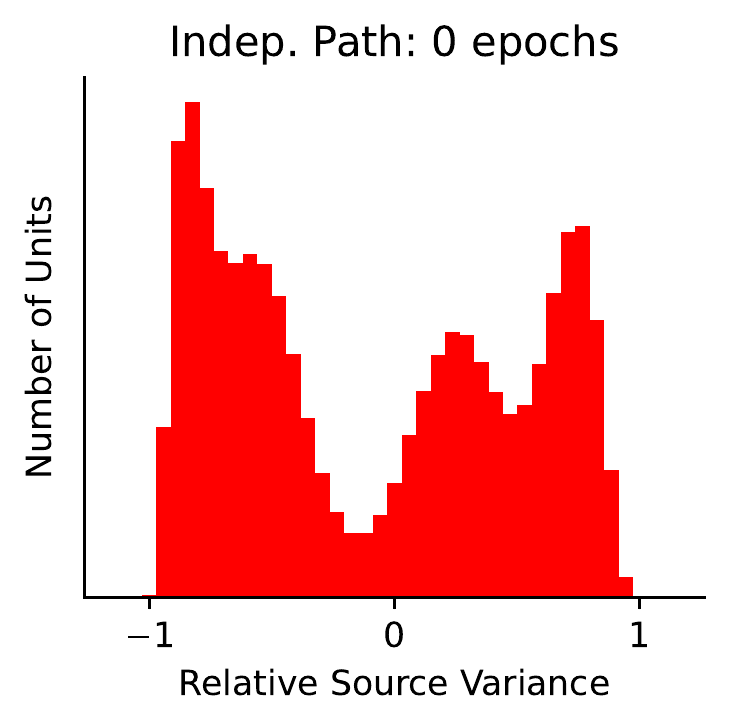}
    \includegraphics[width=0.15\textwidth]{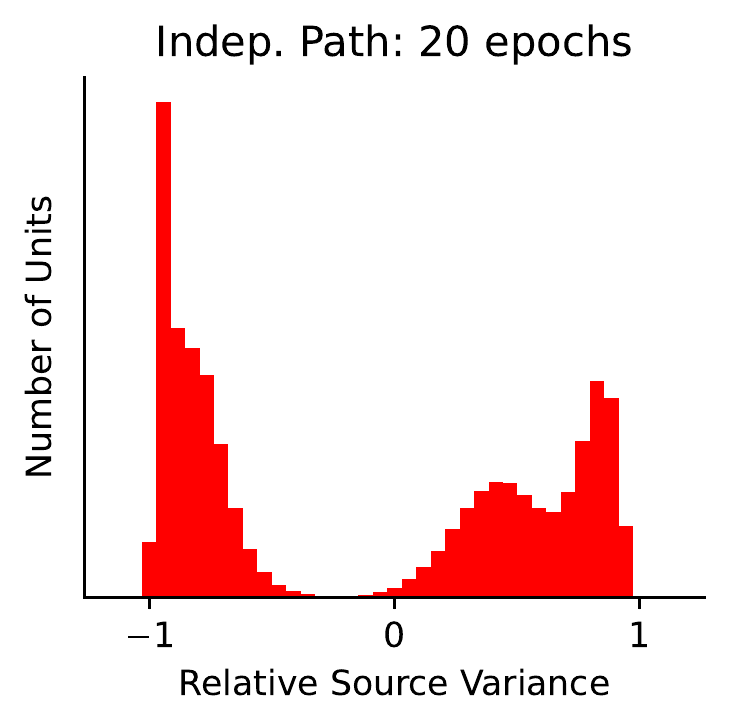}
    \includegraphics[width=0.15\textwidth]{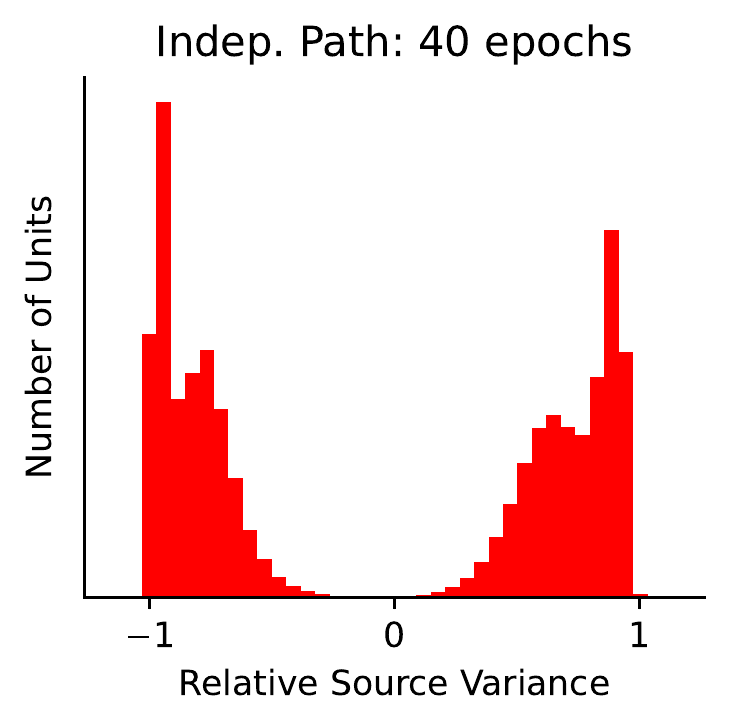}
    \includegraphics[width=0.15\textwidth]{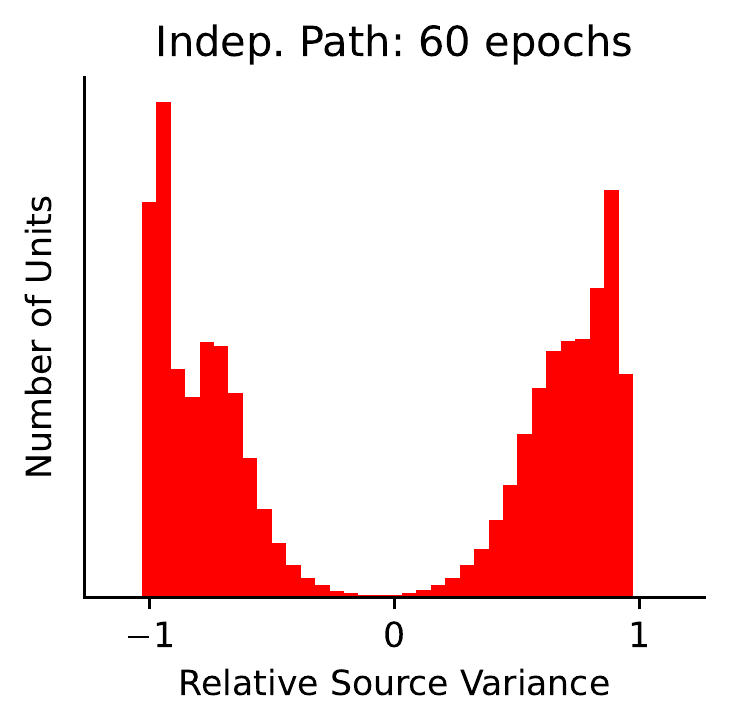}
    \includegraphics[width=0.15\textwidth]{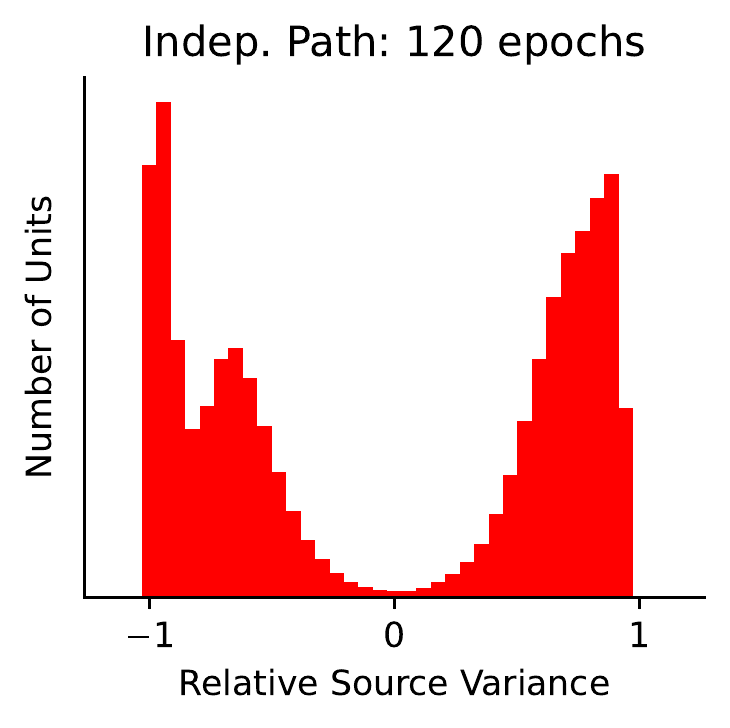}
    \includegraphics[width=0.15\textwidth]{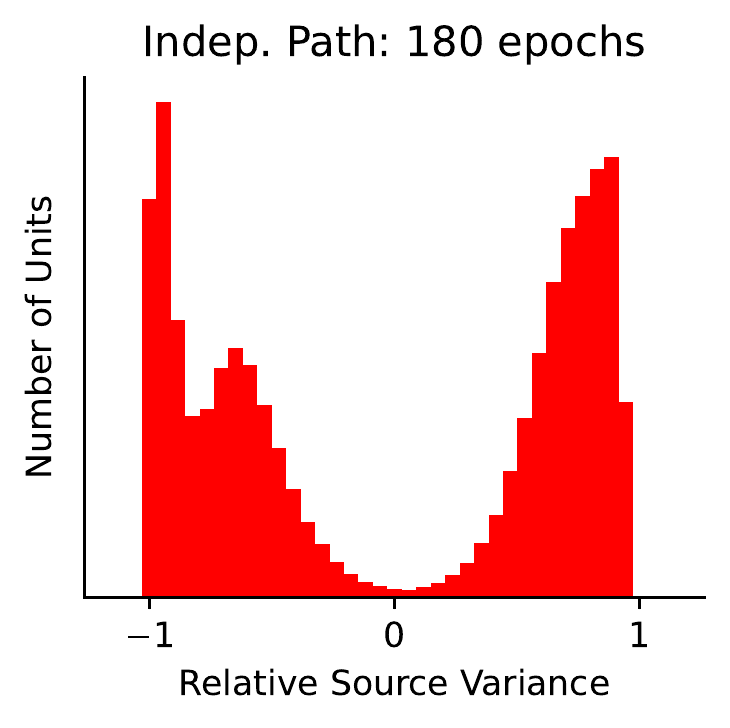}

    \caption{
    Strabismus-Like Deficit for ablation of no weight decay ($\text{wd} = 0$), no data augmentation and initial $\text{lr} = 0.05$. We also observe a polarized representation. Note the performance is reduced in comparison to Fig.~\ref{fig:strabismus}, due to the lack of data augmentation and weight decay.}
    \label{fig:strabismus_noaug_wd0}
\end{figure}

\begin{figure}[]
    \centering
    \includegraphics[width=0.275\textwidth]{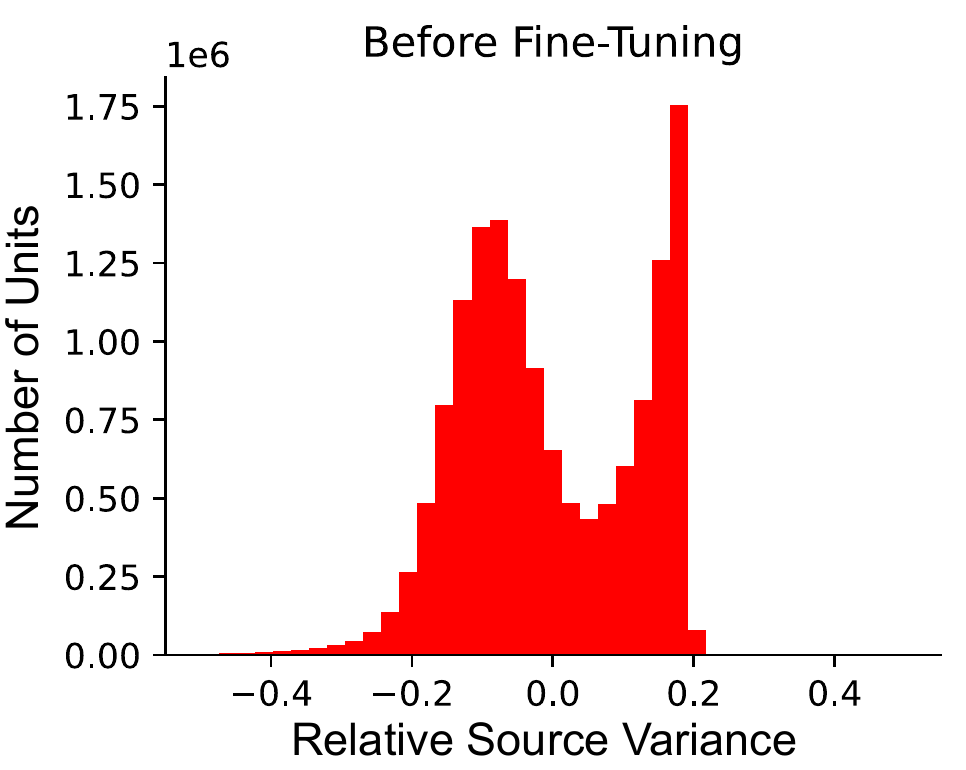}
    \includegraphics[width=0.26\textwidth]{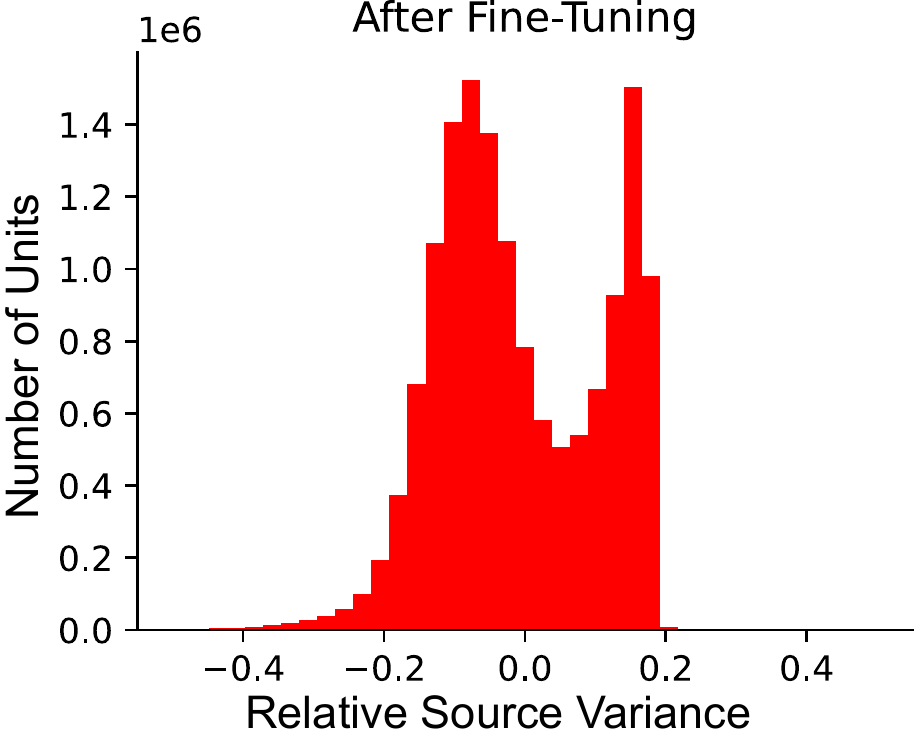}
    \caption{\textbf{Relative Source Variance for Multi-View Transformer.} \textbf{(Left)} We show the distribution of RSV evaluated on the units at output of the encoder before fine-tuning, revealing a bimodal distribution. Here, training was performed without any deficits.
 \textbf{(Right)} During fine-tuning, the representations appear to adapt to become slightly more balanced, depending more evenly on each view, while retaining the initial bimodal structure learned during pre-training.
 \vspace{-1em}
 }
   \label{fig:sensitivity_rsv}
\end{figure}

\begin{figure}[t]
    \centering
    \includegraphics[width=0.3\textwidth]{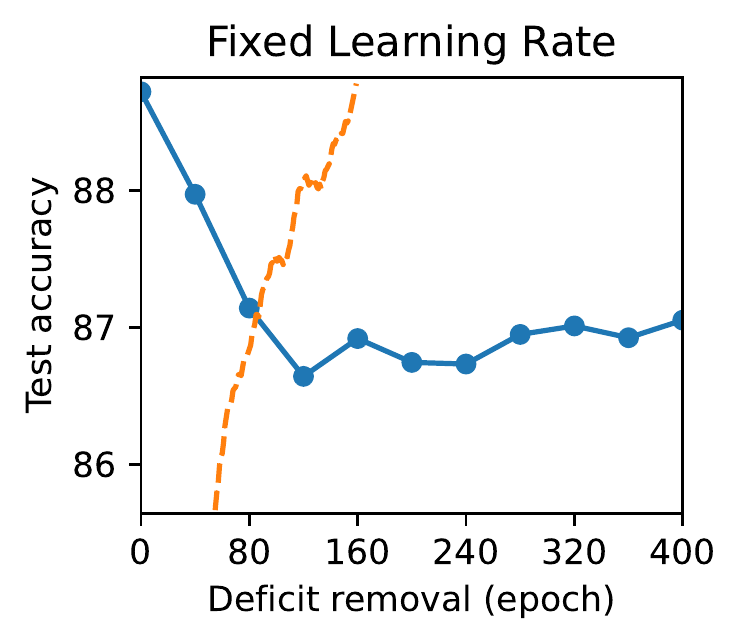} \\
    \includegraphics[width=0.15\textwidth]{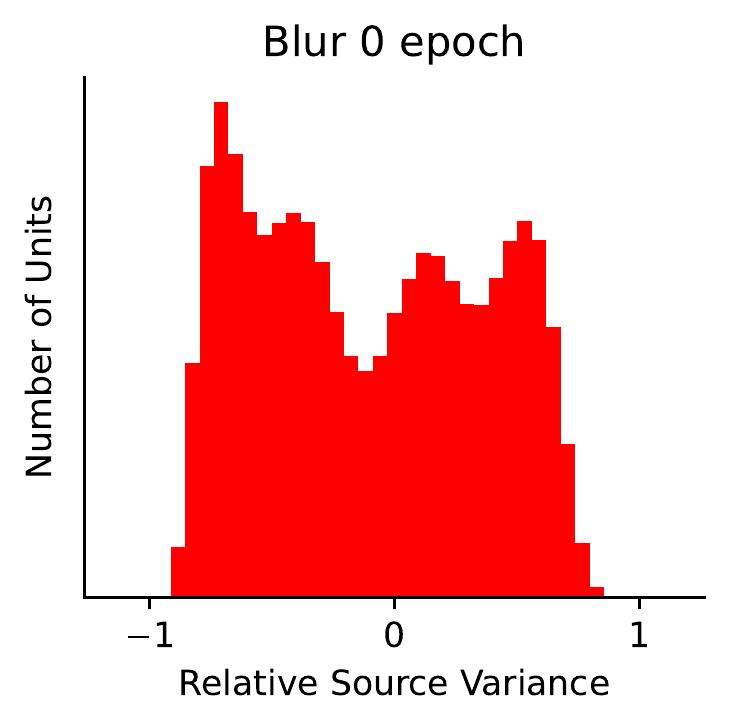}
    \includegraphics[width=0.15\textwidth]{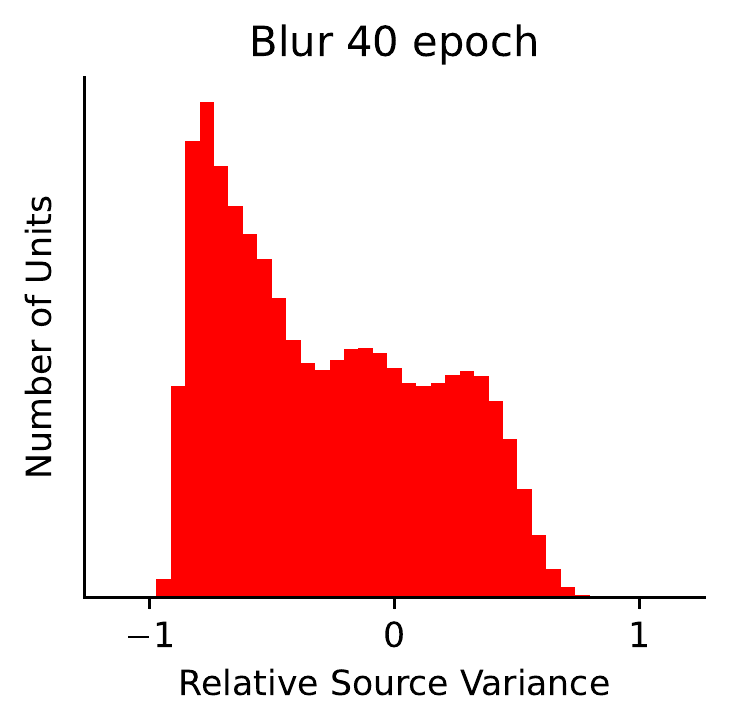}
    \includegraphics[width=0.15\textwidth]{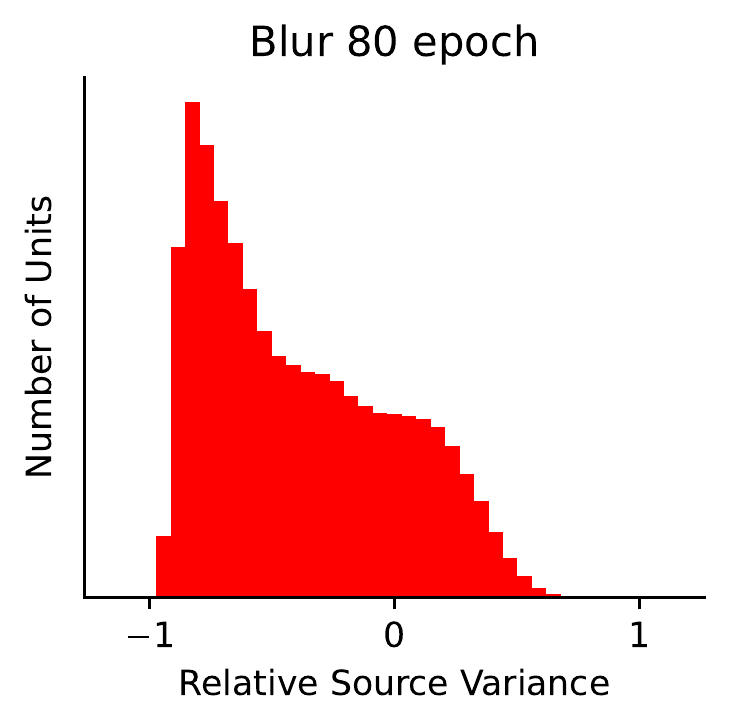}
    \includegraphics[width=0.15\textwidth]{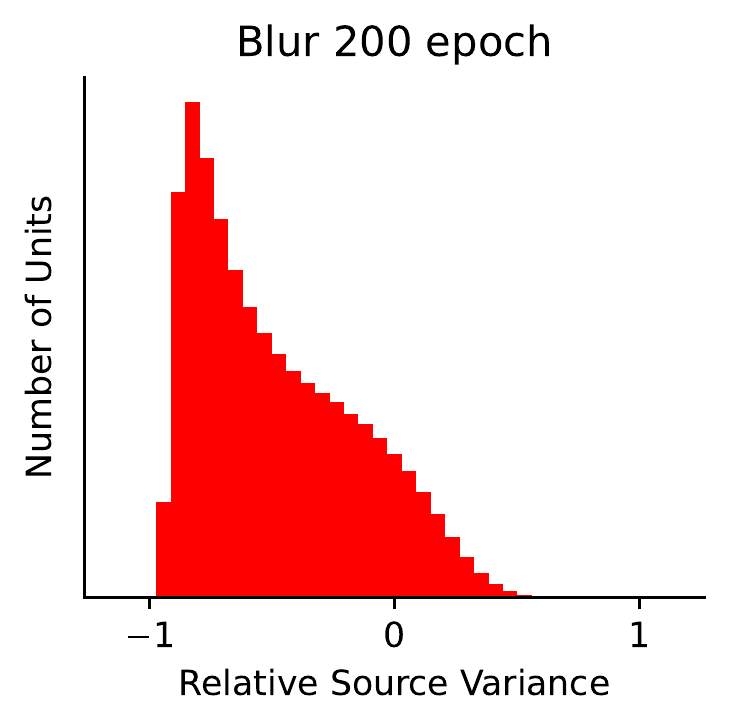}
    \includegraphics[width=0.15\textwidth]{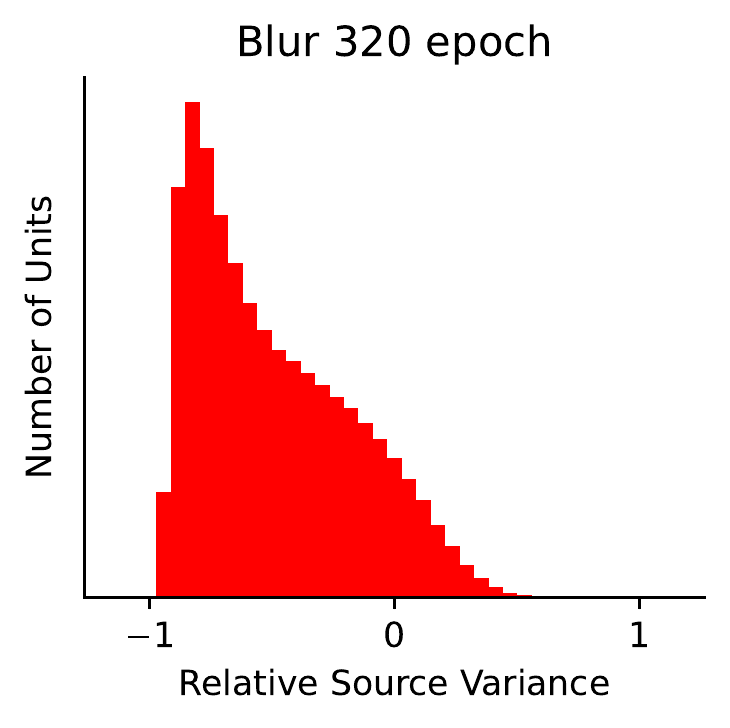}
    \includegraphics[width=0.15\textwidth]{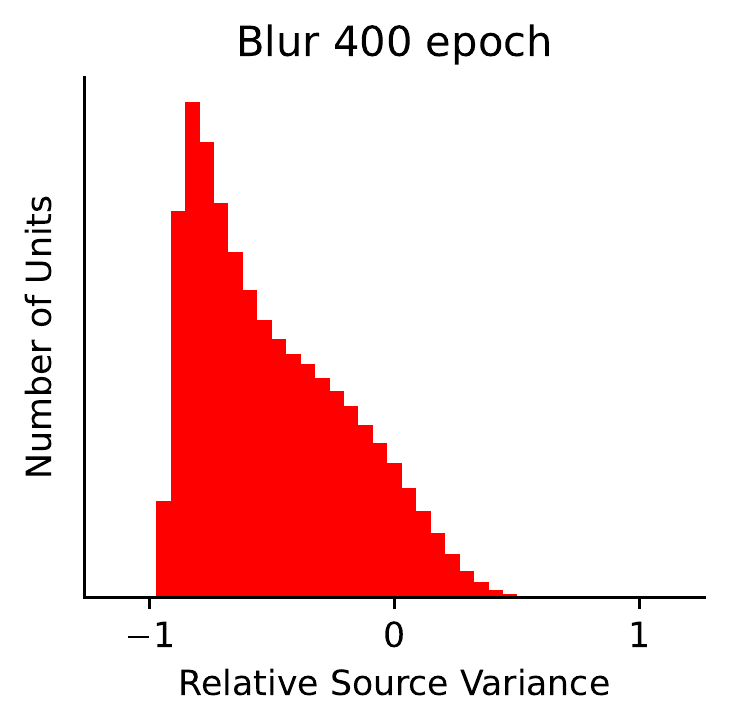}
    \caption{Fixed learning rate of $0.0005$ during training have similarly shaped critical periods to those in paper, and similar RSV distributions as a result of the deficit.}
    \label{fig:fixed-lr}
\end{figure}

\begin{figure}[t]
    \centering
    \includegraphics[width=0.25\textwidth]{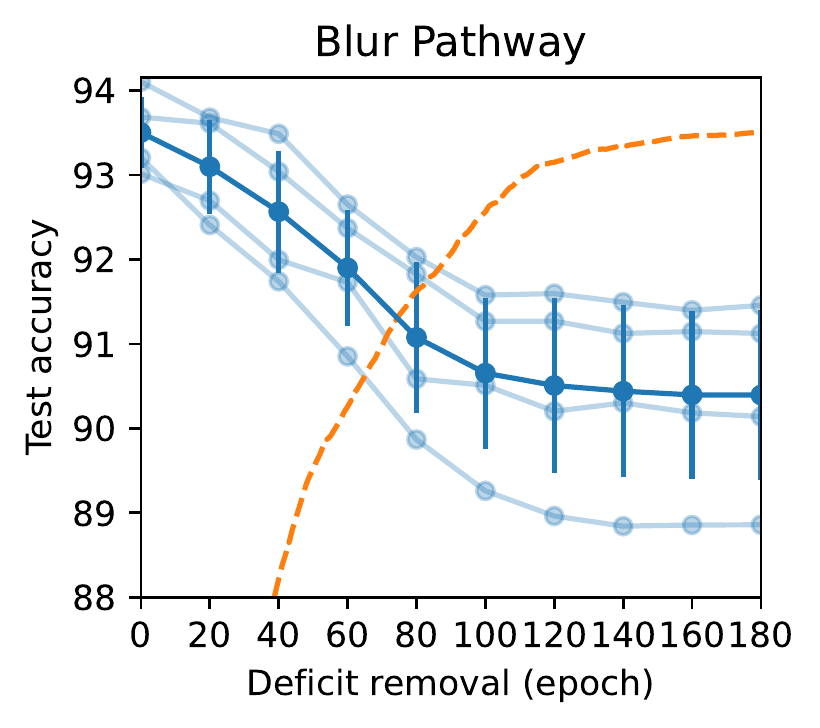}
    \includegraphics[width=0.26\textwidth]{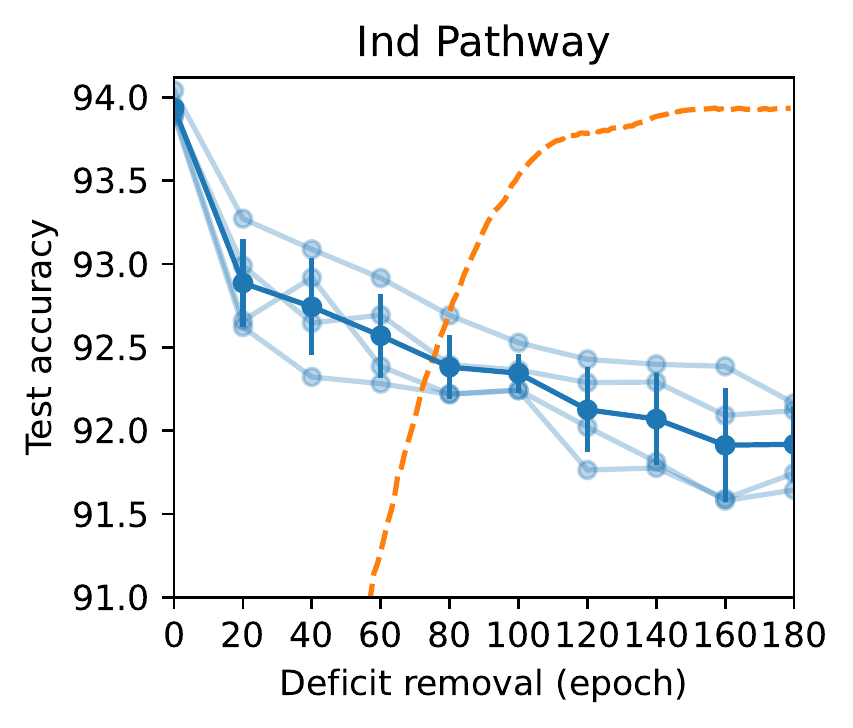}
    \includegraphics[width=0.28\textwidth]{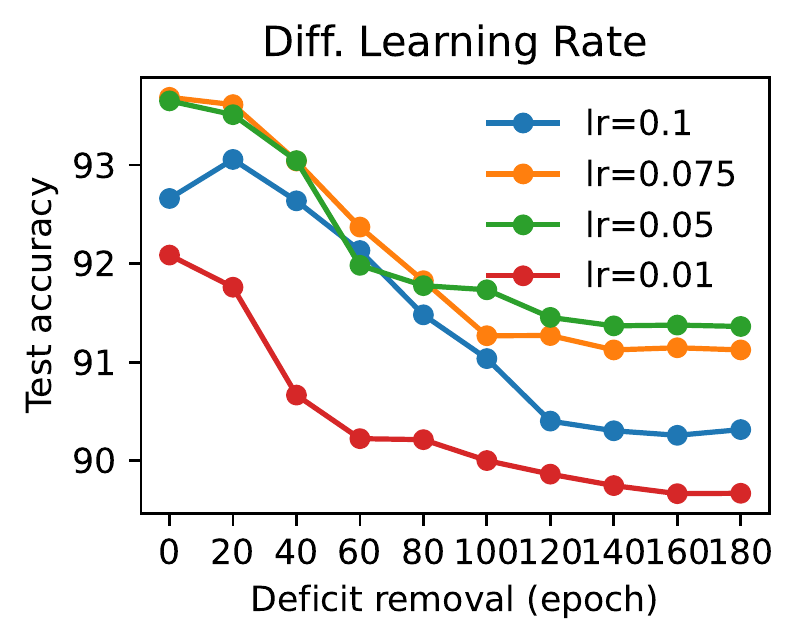}
    \vspace*{-0.3em}
    \caption{Results of multiple runs (light blue), their average (dark blue), and std (bars) for \textbf{(Left)} blurring and \textbf{(Center)} dissociation deficit. \textbf{(Right)} Different initial learning rates (for blur deficit) have have similarly shaped critical periods to those in paper.}
    \label{fig:robustness}
    \vspace{-1.5em}
\end{figure}

\end{document}